**Spreadsheet computing with Finite Domain Constraint Enhancements**

A Thesis

Presented to

The Department of Computer Science

and the

Faculty of the Graduate College

University of Nebraska

In Partial Fulfillment

of the Requirement for the Degree

Master of Science

University of Nebraska, Omaha

By

**Ezana N. Beyenne**

May 2008

# Abstract of the Thesis

## Spreadsheet computing with Finite Domain Constraint Enhancements

Ezana N. Beyenne, Master of Science

University of Nebraska, 2008

**Advisor**: Dr. Hai-feng Guo

Spreadsheet computing is one of the more popular computing methodologies in today's modern society. The spreadsheet application's ease of use and usefulness has enabled non-programmers to perform programming like tasks in a familiar setting modeled after the tabular "pen and paper" approach. However, spreadsheet applications are limited to bookkeeping like tasks due to their single-direction dataflow. This thesis demonstrates an extension of the spreadsheet computing paradigm in overcoming this limitation in order to solve constraint satisfaction problems. We present a framework incorporating a finite constraint solver in a seamless manner with the spreadsheet computing paradigm. This allows the individual cells in the spreadsheet to be attached to either a finite domain or a constraint specifying the relationship that exists among the cells. The framework not only provides an interface for constraint solving, but it also further enhances the spreadsheet computing paradigm by providing a set of spreadsheet-specific constraints that will aid in controlling the scalability of large spreadsheet applications implementations. Finally, we provide examples to demonstrate the usability and usefulness of the extended spreadsheet paradigm.

# Dedications

To my parents, my family members, mentors and friends who have accompanied and inspired me on my journey

# Acknowledgements

I would to thank all the people involved who helped me complete my thesis. First on my list is Dr. Hai-Feng Guo, who has been a phenomenal adviser and teacher. Under his tutelage, I have gained a deeper understanding and appreciation of the research process involved in writing a thesis. The experience of writing this thesis has been wrought with moments of frustrations and triumphs along the way. Dr. Guo has been always been there to listen and provide timely advice throughout the process. I have become a better researcher and student, and for that I am very grateful.

I would also like to thank the rest of my thesis committee member: Dr. Hassan Farhat, and Dr. Song Ci for the insightful suggestions, questions and comments. Their encouragement has made the journey interesting and possible.

Carla Frakes, PKI's outstanding advisor, has also been there to answer any questions regarding starting the thesis to submitting it, in addition to providing advice on my graduate classes. People like her at the college, enable us students to concentrate on our studies, and for that all I can say is thank you.

Tina Ray and Sarfraz Chandio have put up with my countless requests to read my various drafts and provided comments in a timely manner. They also allowed me to practice thesis defense presentations. For their support, encouragement and feedback, I am truly grateful.

Lastly but certainly not least, I would like to thank my parents, family, and friends who continue to inspire me.

# Table of Contents

# Lists of Figures

# CHAPTER 1

# Introduction

## 1.1    Overview

The spreadsheet has been one of the most widely used application in modern society due to its design emphasis on user centered usability and usefulness. End users who are not professional programmers have been able to easily model programming like tasks such as personnel information management, computerized bookkeeping, and electronic financial planning, in a tabular "pen and paper" approach. Each cell in the spreadsheet may contain some data type or a formula which is dependent on values contained in other cells, when these cells have their values updated, the dependent cell is also dynamically updated. This single direction dataflow model has made the spreadsheet very useful since it is easy for users to comprehend, and therefore manipulate. This ease of use and single direction dataflow model at the same time hampers the further application of spreadsheets to solving many daily-life constraint problems, such as resource allocation, task scheduling, and timetabling problems.

Some research efforts have been conducted to further extend the spreadsheet application's ability to solve domain-specific problems by combing it with powerful computational tools.  For example, a sales/products configuration system [14] that had a spreadsheet interface was developed as an Excel add-on, with a lightweight constraint solver known as ILOG Solver. PrediCalc [1] is a spreadsheet application that incorporates general logical constraints for a data management system. These logical constraints are constructed using cell names with the usual logical connectives and quantifiers used to maintain consistency among the databases. Another spreadsheet extension that allows users to account for uncertainties of values by using intervals is the IntervalSolver [6]. This allows the IntervalSolver to deal with rounding errors, imprecise data and numerical constraints on the intervals.

There is other more general extension with other programming paradigms. One of these general extensions is PERPLEX [22] which combines the power of the logic programming with a spreadsheet interface and allows users in an interactive way to define predicates among data cells. Another combination of the programming environment with a spreadsheet interface is the NExSched [12] which uses a finite domain constraint solver (SICStus) as a plug-in for Microsoft Excel to solver constraint satisfaction problems. The Knowledgesheet application [3] a precursor to NExSched, also uses a finite domain constraint solver with java spreadsheet interface to solve constraint satisfaction problems. Another proposed spreadsheet enhancement is NEXCEL [20, 21] which proposes a deductive extension of the spreadsheet paradigm for the purpose of reasoning and decision-making. NEXCEL allows users to define logical inference rules for the purpose of symbolic reasoning, with values satisfying these rules being visualized dynamically and propagated to cells that depend on these rules.

## 1.2  Problem Statement and Approach

All the aforementioned extensions provide a spreadsheet interface for constraint (logic) programming or solving domain-specific problems. Although, these extensions have enhanced the usefulness of spreadsheets without improving their usability, they will remain inaccessible to the general users for whom the spreadsheets were originally designed for. It still requires users to have an understanding of the complex syntax and semantics of the underlying language. This understanding is vital in order for them to define predicates, constraints, and inference rules in order to solve their domain specific problems. This would be a radical departure from the cognitive requirement of the user as a primary design concern of the spreadsheet paradigm [8].

If a user centered design approach is not taken into account, the extensions will be inaccessible to the popular spreadsheet users, and as such their practical contributions will be limited, since domain experts would rather just use the interface of the underlying programming paradigm. For this reason, a user centered design approach to designing the extensions and their function is absolutely crucial. This importance is further echoed in [8] by Jones et al, etc... "*the commercial success of spreadsheets is largely due to the fact that many people find them more* **usable** *than programming languages for programming-like tasks*."

When designing for spreadsheets, we should take into account the reasons for their popularity. Their popularity lies in their ability to provide familiar and concrete representations of data values and formulas in form of functions, while at the same time hiding the syntactic and semantic complexity of traditional programming languages, and its ability to provide immediate feedback.

Another problem that spreadsheet users face is the problem of handling scalability and complexity. Traditional programming languages allow programmers to handle the issue of scalability and complexity through the use of control statements and data structures such as loops, objects and recursion. However the spreadsheet users do not have the assistance of loops and recursion, so they end up tackling quite complex problems by specifying a large number of constraints. This further adds to the complexity therefore making the spreadsheet interface impractical, in addition to making it more error prone for constraint satisfaction problem solving. We provide a set of spreadsheet specific functions to the users to allow them to specify constraints that exist among the cells in a declarative and scalable way.

## 1.3   Contributions and Significance

The main contributions of this thesis are centered around providing a development framework that incorporates the spreadsheet paradigm and constraint programming with finite domain paradigm in such a way that it becomes useful to the general spreadsheet user. This will at the same time enhancing the application's usability.

The usefulness of the development framework lies in the fact that it extends the capabilities of the spreadsheet by providing an interface to a constraint solver. Each cell in the spreadsheet can now be attached to either a finite domain or a constraint in order to specify the interdependent relationship that exists among them. The framework will be responsible for transforming these domains and constraint into a format understood by the backend constraint solver (SWI Prolog), and getting the result(s) back to the spreadsheet in a dynamic and interactive manner.

The general user will be able to use the extended spreadsheet to explore new ways of solving many problems (e.g., scheduling problems, puzzle solving etc..) that previously proved to be an arduous, time consuming task and also inaccessible due to semantic and syntactic complexities . As new ways are discovered by users, the extended spreadsheet will become an even more indispensable tool in solving previously complex and error prone problems that they encounter in daily lives.

The usability of the extended spreadsheet lies in the fact that we provide a new set of spreadsheet specific language functions designed specifically for the general user. This would be the first time, as far as we know that a spreadsheet specific language for constraint solving problems has been designed. This spreadsheet specific language for constraint solving is made up of functions that will allow the users to specify constraints that exist among the cells in a declarative and scalable way. In addition, we provide an interface to allow the experienced constraint programmer to define new predicates. The framework will significantly simplify the complexity involved in developing solutions to many constraint based applications, from within the spreadsheet interface.

A system called CSeE (Constraint Solving enhanced Excel) has been developed to act as an interface tool that exists between Microsoft Excel and SWI Prolog. This work focuses on merging the Constraint Logic Programming (CLP) paradigm with Excel's paradigm in such a way that it hides the complexity associated with CLP (FD). It does this by providing an interactive Add-in to Excel in the form of menus, and spreadsheet specific functions extensions. The menus and spreadsheet-specific functions will enable end users to specify constraints in fewer steps, while providing a dynamic interaction with a back end solver (SWI prolog).

The CSeE system was developed and tested on Windows XP Professional, using Microsoft's .NET framework, the C# language, Visual Studio Tools for Office (VSTO), Excel 2003 and SWI-Prolog. The .NET framework is an ideal platform for developing the CSeE system because it provides access to rich class libraries, templates, technologies and access to features already present in Excel. This will give users a solution that has a familiar look and feel.

Additionally, It allows the manipulation and reusability of Excel's application's object model (e.g. providing interactive menus) and a methodology to communicate with a third party software (e.g. SWI-Prolog) in a seamless manner through the C# language and VSTO.

## 1.4    Organization

The remaining part of this thesis is organized in the following way:  Chapter 2 provides a literary review of some of the relevant research and related works. Chapter 3 gives a brief background of the principles of constraint logic programming that will be required to understand the constraint solving methodologies. Chapter 4 shows the abstract architectural design of the CSeE system, and Chapter 5 provides some of the implementation details of the system. The domain specific examples and results are presented in Chapter 6. Finally, the conclusion and future extensions discussions are presented in Chapter 7, respectively.

# CHAPTER 2

# Spreadsheet Paradigm

The idea of combining the spreadsheet paradigm with a logic engine is not a new one [1, 2, 3, 6, 7, and 12]. In addition to giving an overview of spreadsheet computing, we will give a brief description some of the extensions namely KnowledgeSheet [3], NexSched [12], PrediCalc [1] and XcelLog [2] and our proposed extension known as CSeE.

## 2.1 Spreadsheet computing

Spreadsheet computing has become a popular way for end users to solve real world problems and aid in decision making processes. Due to its interactive nature and user centric design, users without any programming experience can use the application to write their own programs. They do so by just modeling the problem they want to solve as a set of input values consisting of exact numbers to represent real world numerical data and their mutual relationships as functions and the output values are then calculated from the given input values [5]. The problem with the spreadsheet paradigm is that it forces users to use exact values and functions to model real world data which is often incomplete and inexact. The real world is not always exact and answers to some missing constraints can only be determined when the constraints have been pruned, searched and solved [3, 5].

In spreadsheet computing the output relationship that exists among cell values can only be determined when the functions evaluated all the given input cell values [5]. But you cannot have the result and some of the inputs to find the remaining input value, in other words it is not bi-directional.

Finally, in spreadsheet computing only one solution can be derived at a time, and if you want to find other solutions, manual manipulation of the input values is required for each iteration; this is known as the brute force technique [6, 12].

## 2.2 Extensions to Spreadsheet computing

The idea of extending the spreadsheet paradigm with more powerful computational tools for solving domain specific problems is not a new one [1, 2, 3, 6, 7, and 12]. The ones that are at the top of the list are KnowledgeSheet [3], NexSched [12], PrediCalc [1] and XcelLog [2]. Each of these extensions has been able to solve domain specific problems using different approaches.

### 2.2.1 The KnowledgeSheet and the NExSched paradigm

The KnowledgeSheet paradigm is a system that allows users to interactively solve constraint satisfaction problems, by letting the designer attach finite domain constraints to individual cells that are then solved using a CLP (FD) backend solver. It consists of a spreadsheet interface, an third party CLP (FD) engine, which in this case was the ECLIPSe CLP (FD) system and used a protocol that served as a communication bridge between the spreadsheet and the engine [3].

The KnowledgeSheet interface was implemented using the Java language with a limited set of functionality and required that each constraint, domain values and constants be marked with the following tags <C>, <D> and <K> . This method of having to mark each constraint, domain values and constants becomes more cumbersome as the problem domain grows [3].

The KnowledgeSheet system served as a rudimentary interface to a back-end constraint logic system, and only provided native CLP function support, which further limited the user pool to those who had a basic understanding of the constraint logic programming language. Additionally, it provided a spreadsheet interface that lacked most of the functionality of mature spreadsheets such as Microsoft Excel. Therefore the general spreadsheet user would not be able to scale and solve complex problems, since it did not provide a way to handle recursions and loops.

The NExSched system is a system that incorporated the KnowledgeSheet system as an Add-in to Excel using VBA and the SICStus constraint solver [12]. The system relies on macros to facilitate copying of cells/formulas, for built-in functions and arithmetic functions, for solving a problem in one shot or in parts when the problem becomes two large.

Although, the NExSched brought some additional functionality, it did not incorporate a user centric design philosophy, therefore the general user would not find it very practical, and so the audience would be limited to domain experts.

**2.2.2 The PrediCalc and XcelLog paradigm**

The PrediCalc system attempts to extend the capabilities of a spreadsheet with the logic engine, and allows logical relations between cells to be defined, but it is organized in a database like manner. Because of its database like organization of cells into a table like formation, it allows for database like queries of these tables and also the ability to refer to individual cells using names and allows propagation to take place even when there is an inconsistency or just through the use of the structured names of the cells [1].

The XcelLog system is implemented as an Add-in to Excel, and provides a spreadsheet based interface to solving problems with a rule based system. The main users of this system are domain experts who write rule based programs to find solutions to problems by applying rules to given input values. These rule based systems translate the specification to logic programs which are used to deduce the solution.

XcelLog and PrediCalc do not use constraint propagation techniques, instead they make use of logical relations and rely on the ability of the logic engine to deduce and solve the problem. The focus of both these systems was on users who were domain experts; it would remain inaccessible to the popular spreadsheet users.

### 2.2.3 The CSeE paradigm

The CSeE (Constraint Solving enhanced Excel) also extends the spreadsheet computing paradigm, but focuses more on user-centric design characteristics such as the usefulness and usability of the system for the general spreadsheet user. As far as we know this is the first time that the spreadsheet paradigm was extended with the general user in mind, instead of domain specific users.

The CSeE system like the other extensions KnowledgeSheet and NExSched provides a useful interface to a constraint solver. The KnowledgeSheet and NexSched require that users still define predicates and constraints using the correct syntax of the underlying constraint solver. This is where the CSeE system differs from the aforementioned extensions, because we have developed a spreadsheet specific language, which will appear to be familiar to spreadsheet users who are used to working with functions to define relationships among numerical data [5]. These spreadsheet specific functions are functions designed to solve constraint satisfaction problems in fewer steps. This will allow the general users to explore solving problems in new applications areas for which constraint satisfaction techniques are very well suited such as scheduling problems, puzzle solving etc..

KnowledgeSheet requires that domain variables be marked with an HTML/XML looking tag **<D> </D>** (e.g., domain variables located in cells A1, B1, C1 will show up as **<D>**A1**</D>**, **<D>**B1**</D>**, **<D>**C1**</D>**) and constraints be marked with the **<C></C>** tag, as the variable and constraint size grows, this could prove to be very cumbersome and time consuming, thereby introducing the possibility that errors. NExSched uses the key words "*Col Constraint*" and "*Row Constraint*" for constraints and a Map table for domains but spreadsheet users are used to using the functions dialogue, which allows them to drag and highlight a cell or a range of cells, to call functions. CSeE system allows users to call two spreadsheet-specific functions namely: *ssVarRanges*() used to specify variable domain ranges, and *ssConstraintRanges*() used to specify the variable constraint ranges, through the functions dialogue which makes the program look cleaner and easier to change.

CSeE allows users with large and complex problems to separate domains and constraints into various sheets (e.g., domains can be defined in sheet1, sheet2 and constraints can be defined in sheet3). By enabling users to separate domains and constraints over multiple sheets it allows for the clear separation of concerns unlike KnowledgeSheet and NexSched.  CSeE also allows advanced users to create and add to the spreadsheet specific language.

The XcelLog and PrediCalc are not equipped to solve constraint satisfaction problems, and since we are trying to solve constraint satisfaction problems, CSeE is better equipped to solve them using constraint propagation techniques.

# CHAPTER 3

# Constraint Logic Programming

In this chapter, we will give a basic overview of the Constraint Satisfaction problem (CSP), and some of the algorithms used to solve such problems. I will also give brief introduction of the Constraint Logic Programming methodology used to solve CSP problems. For a more in-depth coverage of the Constraint Logic Programming methodology, I will refer you to the reference section, particularly [4] and Dr. Roman Bartak's online guides [13, 15, 16].

## 3.1 Constraint Problems

A constraint is a description of the logical relationship that exists among several unknown variables, where each variable takes values in a given domain [13]. A constraint is therefore used to restrict the possible values that a variable can have [13]. It represents some partially known information about a particular variable. An important characteristic of constraints is their declarative nature, i.e. they do not specify the computational procedure that should be used to enforce the relationships, rather they just state what relationships must hold and leave it up to the solver to find a solution.

There are two main solving technologies involved in finding solutions for satisfying relationships that exist among constraints. These solving methodologies are *Constraint Solving* and *Constraint Satisfaction*.

The *Constraint Solving* methodology deals with constraints that are defined over an infinite and therefore more complex domain. The algorithms used to solve these more complex domains are based on mathematical techniques such as Taylor series, automatic differentiation etc... [13].

The second methodology is *Constraint Satisfaction* and deals with constraints over a finite domain. More than 95% of all industrial constraint applications deal with problems over a finite domain such as scheduling etc [13]. The SWI-Prolog constraint solver used in the CSeE system will be using the *Constraint Satisfaction* methodology.

## 3.2 Constraint Satisfaction Problems

For many years, Constraint Satisfaction Problems (CSPs) have been researched by the Artificial Intelligence community whose motivation lied in finding solutions for difficult combinatorial problems. Due to this research, many powerful algorithms would become the foundation of current constraint satisfaction algorithms. A Constraint Satisfaction Problem basically requires:

i.  A set of variables X={$x_1$... $x_n$}.

ii.  For each variable $x_i$ in the above set, a finite set $D_i$ of possible values. The values do not need to be a numeric in nature, although they often are. If they are numeric values such as integers they do not need to be consecutive integers.

iii. A set of constraints that will restrict the values that each variable can simultaneously have.

A *solution* to a CSP is said to have all constraints satisfied when a value from its domain to every variable is assigned. The desired solution that we maybe looking for:

i. A single solution among many, but with no preference as to which.

ii. All solutions.

iii. A good or optimal solution, when an objective function is defined in terms of all or just a few of the variables.

One might wonder, why modeling a finite domain problem as a CSP might be preferable to other models such as a mathematical programming problem. The reasons are as follows:

1. Problem formulation is simpler and solutions are easier to understood and straightforward since CSPs are much closer to the original problem. The variables of the CSPs correspond to the problem entities, and their constraints do not need to be transformed into their linear inequalities [13].

2. Even though the CSP algorithms are basically very simple, they can occasionally be quicker in finding a solution, than if other methods such as integer programming are used [13].

The solution to a CSP problem are found through the use of systematic search methodologies that traverse through the various assignment values that a variable might have. The traversing can explore the entire space of value assignments to all the variables or just explore partial value assignments to their variables. The next section will discuss the algorithms used to solve CSPs.

## 3.3 Algorithms used to solve CSPs

This section will give a brief discussion on some of the algorithms used to solve constraint satisfaction problems (CSP). Such algorithms consist of systematic search methodologies such as *Generate and Test*, *Backtracking*, and consistency techniques such as *node consistency*, *arc consistency,* and *path consistency*. Finally, some constraint propagation techniques known as *Forward Checking* and *Full Lookahead* will be given a brief introduction.

### 3.3.1 Systematic Search Methodology

Solving CSPs may appear to be trivial from a theoretical perspective due to the simplistic and inefficient nature of the systematic exploration of a search space. The practical point of view differs, due to its emphasis on efficiency. These very simple and inefficient algorithms form a foundation from which optimized algorithms are developed.

One such basic algorithm is the *Generate and Test* algorithm which first generates every possible value assignment for the variables using non-deterministic steps, and then tests the constraints. Generate and Test should only be used as a last resort if all else fails. A more efficient method is the *Backtracking* approach [13].

The *Backtracking* methodology is the most common systematic search method used [13]. This methodology incrementally attempts to get a complete solution, by extending the partial solution by repeatedly choosing a value associated with another variable [13]. This *Backtracking* method has three known major problems namely:

a.  *Thrashing*, repeated failure for the same reason [13]. If we had variables A, B, C, D, and E with their domain ranges from 1...10 and a constraint that states A > E, Backtracking will try all the assignments for B, C, D before realizing that constraint requirement that A > E [16].

b.   *Repeats work already* done since it is non deterministic in nature. For
     example:

     i.   Let us say we have variables A, B, C, D, and E with their domain
          ranges from 1...10 and constraints that state B+8 < D, and C= 5*E.
          During the labeling stage for variables C, E *Backtracking* will
          repeatedly check the values 1…9 for the variable D [16].

c.   *It is incapable of detecting a conflict until it occurs* [13]. For example:

     i.   Let us say we have variables A, B, C, D, and E with their domain
          ranges from 1...10 and a constraint A=3*E. *Backtracking* will find
          out that A > 2 only when it is in the labeling stage for E [16].

Since *Generate and Test*, and *Backtracking* search methodologies have a disadvantage of
relying on a late detection of inconsistencies, consistency techniques were introduced to
prune the search space.

### 3.3.2 Consistency Techniques

Consistency techniques were introduced to alleviate the problem that GT and BT
have of late detection of inconsistent value. Consistency techniques remove inconsistent
values as soon as possible by extending any partial solution of a smaller sub network to
some other network [13]. These algorithms are the *node-consistency* technique, the *arc-consistency* techniques, and *path consistency*.

CSPs are usually represented as a constraint network graph, where the nodes are
the variables and the edges the constraints [13]. Therefore, this requires the descriptions
of the problems are in a special form known as binary CSP. Each binary CSP is either a unary or binary constraint e.g. X ≠ Y, Y ≠ Z, X ≠ Z, X < 5. Arbitrary CSPs can be transformed into their equivalent binary CSP. Binarization in practice can turn out to be not worth the trouble, since the algorithms can be extended to handle non-binary CSPs [13].

The *node consistency* technique basically removes values from the domain of a variable that is not consistent with the unary constraint. We say that a constraint satisfaction problem is node consistent if and only if all the values for all the variables satisfy the constraints on that variable [13]. Fig 3.1 have three variables A, B, C with domain ranges from 0…9 and constraints that state A > 5, A≠ B, A< C, B = C. When one applies the node consistency technique, variable A's range of values changes to 6,7,8,9 in order to satisfy the constraints.

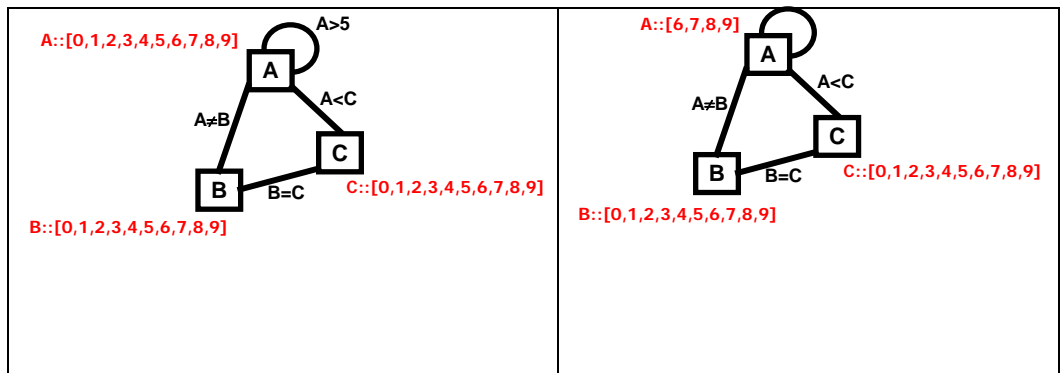

**Fig 3.1** Node Inconsistency to Node Consistency

The *arc consistency* technique removes values from the domain of a variable that is inconsistent with its binary constraints. Assuming we are now dealing with binary constraint satisfaction problems, we can say that an arc (edge) corresponds to a constraint in the constraint network [16]. Therefore, if we had an arc $(Vi, V_{j)}$, it is consistent if and only if for each value *x* in $V_i$'s current domain which satisfies $V_i$'s constraints, there is a corresponding value *y* in $V_j$'s current domain which satisfies $V_j$'s constraints, such that the valuation $V_i=x$ and $V_j=y$ is allowed by the binary constraint that exists between $V_i$ and $V_j$ [13]. A constraint satisfaction problem (CSP) is said to be arc consistent if both arcs $(V_i, V_j)$ and $(V_j, V_i)$ are arc consistent [16]. The Fig 3.2 show how three variables X, Y, Z with domains 1.2 and constraints X = Y, X ≠ Z, and Y > Z, become arc consistent with the value of X and Y becoming 2, and Z's value becoming 1.

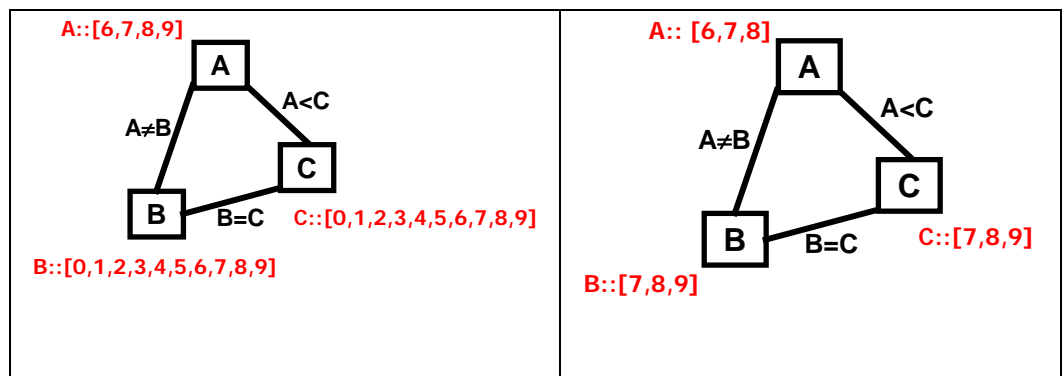

**Fig 3.2** Arc Inconsistency to Arc Consistency

There are several algorithms that use the *arc consistency* technique ranging from AC-1 to AC-7. These algorithms repeatedly perform revisions of the arcs until either there is no solution, in which case the domain is empty or a consistent state is reached [13].

Even more values can be removed using another consistency technique known as the *path consistency* technique is used. This basically states that any pair of assignments can be extended to a third [13].

### 3.3.3 Propagation Techniques

The aforementioned methodologies and techniques can each be used to individually solve CSPs, but this is not frequently done [13]. Combining both approaches are more commonly used to solve CSPs. Although there are many propagation techniques that are employed, we focus on the *Lookahead propagation* technique. *Lookahead propagation* techniques were developed to help prevent future conflicts by applying consistency checks among those soon to be instantiated variables. In other words, *Lookahead* checks to see if there are any inconsistent values in the domains of other variables, when a change to the domain of an argument occurs. We will give a brief description two Lookahead propagation techniques known as *Forward Checking (FC)* and *Full Lookahead (LA)* methodologies.

The *Forward Checking* methodology is the easiest of the Lookahead strategies [13]. It simply applies the arc-consistency technique between pairs of instantiated and uninstantiated variables i.e., when a current variable gets assigned any value, it temporarily removes any value in the future variable's domain that conflicts with the current assignment [4, 13, 15].

The *(Full) Lookahead methodology* does more work than *Forward Checking (FC)* in preventing future conflicts, by extending consistency checks to variables that are non-directly connected [13]. This is in addition to applying arc consistency algorithms to the direct neighborhood just like *Forward Checking (FC)*. Fig 3.3 shows how the Lookahead strategy extends the Forward Checking methodology.

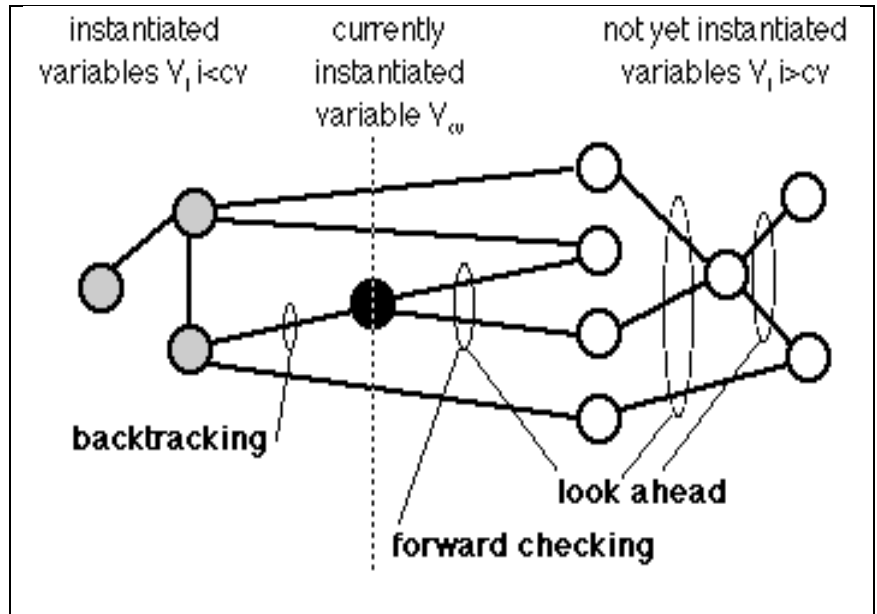

**Fig 3.3** Forward Checking and Lookahead methodologies

The figure above shows that Lookahead methodology does even more work and in some cases can prove to be more expensive than either Forward Checking (FC) or Backtracking (BT) algorithms [13]. So FC and BT are therefore the still used in lieu of Lookahead [13].

## 3.4 Constraint Logic Programming

Constraint logic programming (CLP) is a result of integrating concepts from constraint programming (which deal with constraints over a finite domain) into logic programming.

The main advantage of logic programming is its declarative nature, in which a user states what needs to be solved rather than how it needs to be solved [13]. This is similar to the concept of constraint programming. Logic programming languages such a Prolog have proven to be inefficient when there is a large problem search space. The inefficiency of Logic programs is due to their use of the *Generate and Test* technique which is an algorithm which applies the test *posteriori*, to detect inconsistent values for variables [3].

Unlike logic programming languages, constraint programming is able to have a smaller search space through the use of constraint satisfaction technique known as consistency techniques. By using consistency techniques, the pruning of the search space occurs a *priori* which removes incompatible values found in the domain of variables. The constraints that cannot be solved a *priori* are held in a suspended state until enough values are found to give a solution. Essentially, Constraint programming methodology uses a *test and generate* technique, unlike logic languages which use *generate*, and then *test* [3].

Constraints languages can be embedded within programming languages such as Prolog, or as separate software libraries. There are several constraint logic programming compilers that exist today ranging from open source to proprietary systems, namely GNU Prolog, B-Prolog, ECLIPSe (proprietary), and SWI Prolog (open source) [19].

### 3.4.1 Constraint Logic Programming Examples

A constraint logic program with Finite domains CLP (FD) consists of three parts. They are as follows:

1. *Variable generation* which is responsible for generating variables and also specifying their domains.

2. *Constraint generation* is responsible for specifying constraints over the variables in their domain. Once the constraints are specified over the variables consistency and propagation techniques are specified to reduce the domain.

3. *Labeling* enumerates through the remaining the variables further reducing the domain also using consistency and propagation techniques. This methodology can be used to find the optimized solution to a constraint, since this is used to generate all the possible concrete values of the variables [4]. Since labeling can generate a huge number of possible answers, the labeling call is delayed until the last possible minute [4].

Examples of CLP (FD) programs and solutions are shown below:-

**Example 1: The Knapsack Problem**

A smuggler has a knapsack that has limited capacity of only 9 units. He can smuggle bottles of whiskey of consisting of 4 units, perfume bottles of 3 units and 2 unit cartons of cigarettes. The profit from smuggling a bottle of whiskey is 15 dollars, while a bottle of perfume is 10 dollars or a carton of cigarettes is 7 dollars. The smuggler can only make the trip if he can make a profit of greater than or equal to 30 dollars. What are the combinations that will maximize his profit?

**Solution:**

You can build the solution by defining the problems as follows:

1) This stage is the *variable generation* phase. In this case the variable W will be for whiskey, P is perfume, and C equals cigarettes. Their domain will range from 0 to signify that those items will not be taken, to the max capacity of the knapsack which in this case will be 9. This is done by calling the *fd_domain* function, which will assign the 3 variables values from 0 to 9.

2) The next step is the constraint generation phase. The 4*W + 3*P + 2* C #=< 9 line states that you can carry a combination of 4 units of whiskey , 3 units of perfume and 2 units of cigarettes but the combinations must not exceed the maximum carrying capacity of the knapsack which is 9. The 15 * W+ 10*P + 7* C #>= 30 states that the combination of items he carries must be greater than or equal to 30.

3) The last step is the labeling. For this we use the function fd_labeling (([W, P, C]). The optimized answer will show that the maximum profit can be derived by applying an optimization function such as minimum, maximum to the labeling process. Since we want to maximize the profit, a maximum function will be applied to the labeling function on line 5 in the code below as such:

$$maximize \text{ (labeling([W, P,C])).}$$

One of the resulting optimal answers will be (W) = 1, (P) = 1 and (C) = 1.

```
Knapsack ([W, P, C]):-
              domain([W, P,C],0,9),          (1)
              4*W + 3*P + 2* C #=< 9,         (2)
              15 * W+ 10*P + 7* C #>= 30,     (3)
              labeling ([W, P, C]).           (4)
```

## Example 2: A Precedence Scheduling Problem

The precedence scheduling problem [17] involves four tasks namely *a*, *b*, *c* and *d* that have the durations 2, 3, 5 and 4 hours. There are precedence constraints that exist among the tasks: task *a* has to be before task *b* and *c*, while task *b* has to precede task *d*. Find the start times *Ta*, *Tb*, *Tc* and *Td* of the tasks so that the finishing time *Tf* is minimized. The start time is assumed to be 0. The CSP problem can be stated as follows:-

*Variables*: Ta, Tb, Tc, Td, Tf

*Domains*: all the non-negative real numbers

*Constraints*:

Ta + 2 <= Tb (task a which takes a 2 hours precedes b)

Ta + 2 <= Tc (a precedes c)

Tb + 3 <= Td (b precedes d)

Tc + 5 <= Tf (c finished by Tf)

Td + 4 <= Tf (d finished by Tf)

*Criteria*: minimize Tf

The CLP (FD) Program for the above scheduling problem will generate an optimal solution *StartTimes = [0, 2, 2, 5, 9]* for the code shown below.

| | |
|---|---|
| StartTimes = [Ta, Tb, Tc, Td, Tf], | *% Tf finishing time* (1) |
| domain (StartTimes, 0, 20), | (2) |
| Ta #>= 0, | (3) |
| Ta + 2 #=< Tb, | (4) |
| Ta + 2#=< Tc, | (5) |
| Tb + 3 #=< Td, | (6) |
| Tc + 5 #=< Tf, | (7) |
| Td + 4 #=< Tf, | (8) |
| minimize (labeling ([], StartTimes),Tf). | (9) |

# CHAPTER 4

## Design

This chapter gives the design description of the CSeE system, with the implementation details being discussed in the next chapter. The system consists of four major components that interact with each other in a bi-directional manner. The overall design is shown in the Fig 4.1 below.

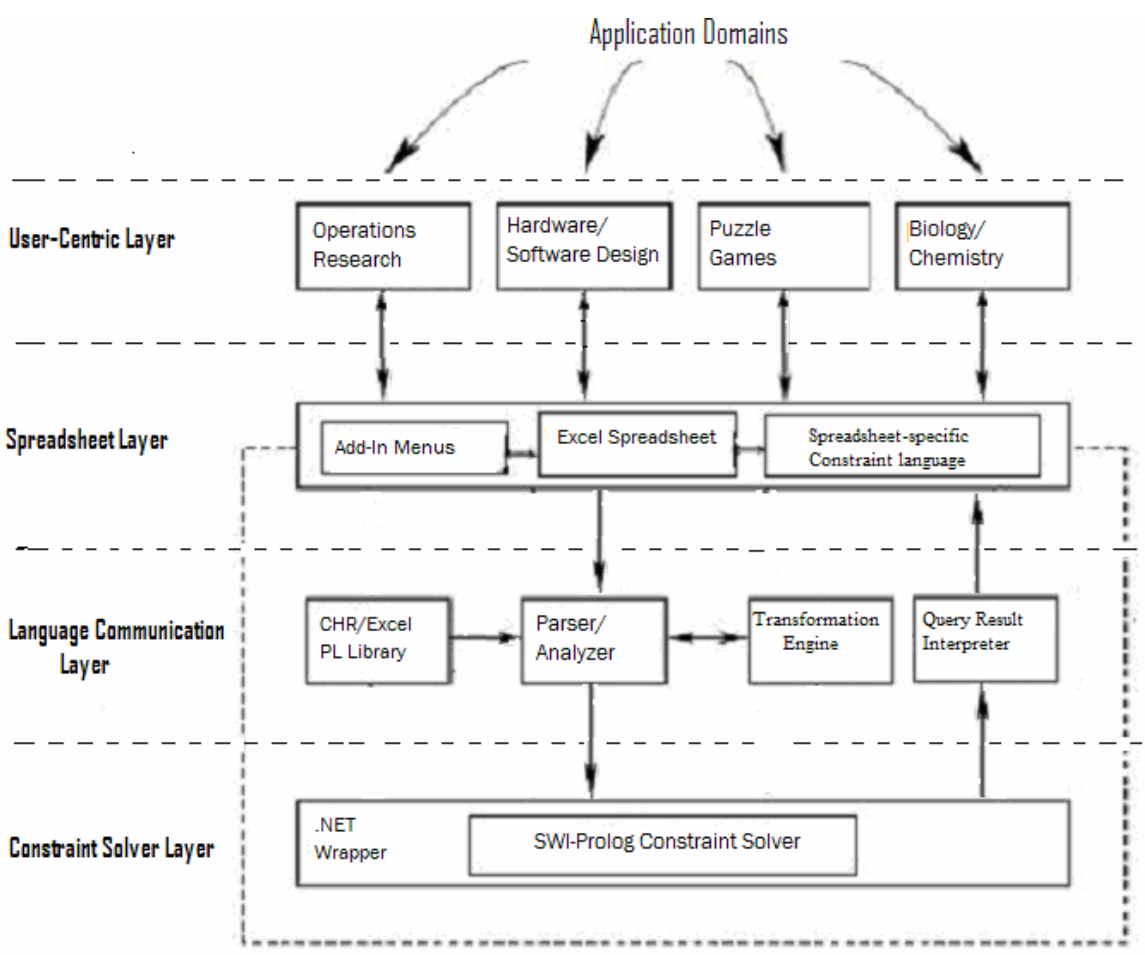

Fig 4.1: CSeE System Overall Design

The CSeE's system's main purpose is to provide a framework that will aid in controlling the scalability of solving large scale CSP problems, through the use of spreadsheet-specific constraint functions. Additionally, this framework provides a seamless integration of the spreadsheet interface with a finite constraint solver, by allowing the individual cells of the spreadsheet to be attached to either a finite domain or a constraint specifying the relationship that exists among the cells.

The *user-centric application layer* of the framework focuses on the user's ability to model and program a large class of constraint satisfaction problems (CSP), in the same manner that they use ordinary spreadsheets to program a class of problems that involve arithmetic computation. These classes of constraint satisfaction problems consist of domains, as diverse as hardware design (e.g. circuit analysis and verification), software design (e.g. protocol simulation and testing), puzzle game solving (e.g. Sudoku), operations research (e.g. scheduling), biology (e.g. protein docking), and finance (e.g. option trading).

The *spreadsheet layer* which lies between the *user-centric application* and *language communication* layers is responsible for capturing and satisfying the user's requests such as button clicks, displaying spreadsheet specific constraint functions etc... It provides menus and access to spreadsheet-specific language functions to allow users to model and program constraint satisfaction problems in a scalable manner. It is also responsible for sending modeled CSP problems to the *language communication layer*, receiving the results in the appropriate format.

The *language communication layer* receives the program information from the spreadsheet layer and performs some analysis on the provided information such checking for required functions. It then proceeds to parse and build the information into a constraint logic program. Once this constraint logic program (CLP) is built, it will send the CLP program to the constraint solver layer, and get feedback on whether it was successfully compiled or if there were errors. Additionally, it is responsible for sending a CLP query through the parser and getting results back from the constraint solver layer and formatting it to a spreadsheet friendly format using the query result interpreter.

The *constraint solver layer* is comprised of the .NET wrapper that acts as the go between for the language communication layer and the constraint solver layer. The .NET wrapper allows the CSeE System to interact will the constraint solver in a .NET managed environment. The constraint solver will parse the generated CLP file and with the aid of the communication wrapper send the results of parse to the language communication layer.

The framework is designed as a set of layers that can be removed, added to or enhanced, without having to do a lot of reworking or redesigning. This will significantly simplify the complexity involved in developing scalable solutions to many CSP problems, from within the spreadsheet interface. The next sections will discuss the design layers in more depth.

# 4.1 User Centric Application Layer

Spreadsheets are a popular computing paradigm due to their ease of use and usefulness that has enabled non-programmers to perform programming like tasks in a familiar tabular "pen and paper" setting. The CSeE framework's user-centric layer provides familiar and concrete representations of domain variables and their values, as well as constraints on these domains in the form of functions. It also provides the added benefit of hiding the syntactic and semantic complexity of constraint logic programming languages while at the same time providing immediate feedback to the user.

The spreadsheet's emphasis on user centered usability and usefulness have enabled end-users to easily apply its tabular "pen and paper" like approach in solving tasks like personal information management, bookkeeping and electronic financial planning. But this ease of use and single direction dataflow model at the same time hampers the further application of spreadsheets to solving many daily encountered constraint problems such as resource allocation, task scheduling, puzzle solving and timetabling problems. The CSeE framework provides the necessary tools to aid in modeling and solving some of the constraint problems. These tools will be demonstrated in domain application areas such as puzzle games i.e. Sudoku, 8-Queen, and resource allocation problems.

The framework also enables spreadsheet users to handle scalability and complexity. Traditional constraint programming languages provide control statements and data structures such as loops, recursions and objects. Without the aid of control statements and data structures, spreadsheet users will find that as the scale of the problem increases, they would have to solve the constraint satisfaction problems using a larger number of constraints, until the problem becomes unmanageable and impractical. The framework provides a set of spreadsheet specific constraint functions that would allow users to specify constraints that exist among cells in a declarative and scalable way.

Spreadsheets lack the ability to define re-usable abstractions (functions), therefore from a programming language point of view they deny "end users" the most fundamental way to control the complexity. Advanced users will be able to add new spreadsheet specific functions to the language library. Therefore the CSeE framework will go from just being an interface to an actual system, which can be used and enhanced by users.

## 4.2 Spreadsheet Layer

The *spreadsheet layer* provides the user with menus for activating desired events that interactively send and receive actions with the other layers such as the language communication layer and constraint solver layer. This layer also provides users access to the spreadsheet specific language functions discussed in the previous section that will enable them to build programs to solve constraint satisfaction problems.

### 4.2.1 Add-in Menus and Sub-Menus

The CSeE's main interface with the *user-centric application layer* is through the spreadsheet layer. It provides a set of menus and sub-menus, which are responsible for capturing specific user events and sending it to the *language communication layer*. Fig 4.2 shows the menus and sub-menus that are provided to capture the parse/build events, and solution displaying options.

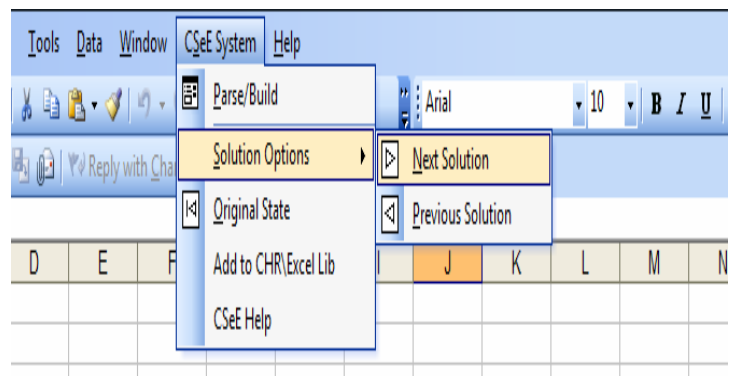

Fig 4.2 CSeE System's menu and sub-menus

The "Parse/Build" menu activates the event to check if the domains variables and constraints on these variables have been specified. It then proceeds to send the gathered information to the parser/analyzer, which build the CLP programs or provides feedback if there were any errors found. If there are no errors found it send a CLP query, gets the number of solutions, and displays them in the "Solutions Options" menu's square brackets [#].

The "Solutions Options" menu lets users to interactively view the available solutions through its sub-menus "Next Solution" and "Previous Solution". The "Next Solution" sub-menu allows the end-user to iteratively view the next solutions, when available. The "Previous Solution" sub-menu provides the users with a way to go back to the previous solution. The "Original State" menu provides users with a way to reset the spreadsheet back to the original state.

The "Add to CHR\Excel lib" menu allows advanced Constraint Logic Programming users to add new Prolog predicates to the Excel pl library. This will allow end users to solve more constraint satisfaction problems. The users will be able to apply and call those functions directly on the spreadsheet, and during the parsing process the new predicated called directly by the parsed CLP program. The "CSeE Help" menu provides a listing of the spreadsheet specific functions and examples of their uses.

The system also provides an interface so that users can use the familiar Excel formulas to select the spreadsheet-specific functions designated with the *ss* prefix, and another so that they can set the selected function's arguments as shown in Fig 4.3.

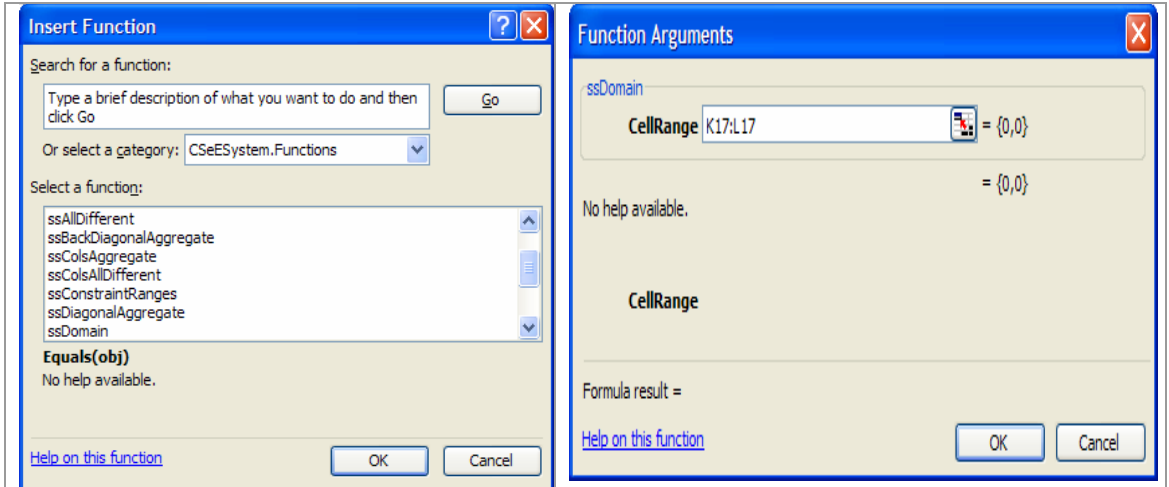

Fig 4.3 CSeE System's insert function and set function arguments interfaces

## 4.2.2 Spreadsheet-Specific Constraint Functions

Each cell in a spreadsheet can be made to attach to a finite domain, or a constraint describing the relationship that exists among the cells and also optimization objectives.

We will be describing the spreadsheet-specific functions as follows, each of which will have a prefix of *ss*. We will describe the syntax, and the predicates that the CSeE system's framework will provide. The syntactic conventions associated with spreadsheet layer in regards to the cell range and values are shown below:

The CSeE framework relies on some of Excel's *A1 reference style* [21], which refers to columns starting with letters beginning from A through IV ( a total of 256 columns) and refers to rows consisting of numbers from 1 through 55536.  The letters and numbers together are the row and column headings. In order to refer to a single cell, you would enter the column heading letter immediately followed by the row number e.g. A1, refers to the cell in column A, row 1. The cell ranges ***CellRange*** can be referenced using the syntax in Fig 4.4 below.

| Use | To refer to |
|---|---|
| A1 | A single cell domain variable, in column A and row 1. |
| A1:B2 | the range of cells in columns A through B and rows 1 through 2. |
| A5:A30 | the range of cells in column A and rows 5 through 30 |
| [A1, A5, B6] | is used to specify the enumeration of variables separated by commas. |
| Sheet2!B1:B10 | the name of the worksheet and an exclamation point (!) precedes the range reference. So **Sheet2** is the name of the worksheet, **!** separates the sheet reference from the cell reference B1:B10. |
| $C_{TL}:C_{BR}$ | is the notation that will be used to the refer to the range of cells starting from the top left through to the bottom right corner. It similar to the A1:B2 reference style with referring $C_{TL}$ to A1 and $C_{BR}$ referring to the bottom right B2. |

Fig 4.4 **CellRange** syntax and meaning

The Domain of the variables will consist of sets of integer values and can be written as:

[Integer$_1$,…, Integer$_k$]         a set of given k finite integers

Int1..Int2                  a set of finite integers X where Int1 $\leq$ X $\leq$ Int2

The arithmetic constraints will have the form of:

*Expression1* **RelOp** *Expression2*

where Expression1 and Expression2 are arithmetic and **RelOp** is a relation as shown in

Fig 4.5 below:

| Relation | Meaning |
|---|---|
| #= | equal |
| #\= | not equal |
| #< | less than |
| #> | greater than |
| #=< | less or equal |
| #>= | greater or equal |

Fig 4.5 **RelOp** syntax and meaning

The relational operators above are different from those used by the Excel spreadsheet, and is adopted from the SWI-Prolog constraint solver.

***BinArithOp*** is the symbol that stands for binary arithmetic operators namely:-

| BinArithOp | meaning |
|---|---|
| + | Addition |
| - | Subtraction |
| * | Multiplication |
| / | Division, unless explicitly stated this operator is not supported. |

Additional arithmetic operators that are supported for *Expressions* described above include the following:

| Additional Operators | meaning |
|---|---|
| **max**(Expression1,Expression2) | The maximum of two expressions |
| **min**(Expression1,Expression2) | The minimum of two expressions |
| Epression1 **mod** Expression2 | Expression1 modulo Expression2 |
| **abs**(Expression1) | The absolute value of an expression |

## 4.2.2.1 Domain Specification Functions

The spreadsheet cells finite domain can be specified individually or as a range of variables as follows:

**[0, 1, 2, 3]**:  Integer domain values can be attached to an individual cell whose value can only evaluate to the integer values in between the square brackets. For example:

|  | A |
|---|---|
| 1 | **[0, 1, 2, 3]** |

**Min..Max**:  Integer domain values can be attached to an individual cell in the form of

Min..Max where Min represents the lower boundary and Max represent the upper boundary of values. For example:

|   | A |
|---|---|
| 1 | **1..3** |

**ssDomain (*CellRange*, Min, Max):**  means that all the variables in the *CellRange* described above will have domains Min..Max as shown in the examples below:

*Example 1*:

*ssDomain(A1:B2,1,3)* it is equivalent to individually assigning domains of 1..3 to cells A1, A2, B1 and B2.

|   | A | B |
|---|---|---|
| 1 | **1..3** | **1..3** |
| 2 | **1..3** | **1..3** |

*Example 2*:

*ssDomain([A1, A2, B2],1,3)* it is equivalent to individually assigning domains of 1..3 to cells A1, A2, and B2.

|   | A | B |
|---|---|---|
| 1 | **1..3** |  |
| 2 | **1..3** | **1..3** |

## 4.2.2.2 Constraint Specifications

The cells in the spreadsheet can as well be used to attach the constraint relationship that exists among the cells that contain the finite domain variables. The spreadsheet specific functions with *CellRange* specified can handle the cell and range references described above, otherwise the specific cell and range references will be required.

**ssAllDifferent (*CellRange*)** : This means that all the variables in the ***CellRange*** will have different values. In the examples below, the table represents the use of the domain variable setting function *ssDomain*, and its equivalent cell range setting. The ssAllDifferent is shown after that.

*ssDomain(A1:B2,1,4)* is equivalent to setting the cell range variables A1:B2's values in the table below:

|   | A | B |
|---|---|---|
| 1 | **1..4** | **1..4** |
| 2 | **1..4** | **1..4** |

*Example 1*:

The *ssAllDifferent(A1:B2)*, is equivalent to defining the constraints in the pseudo code manner below:

$$A1 \neq A2, A1 \neq B1, A1 \neq B2,$$
$$A2 \neq B1, A2 \neq B2, B1 \neq B2$$

*Example 2*:

The *ssAllDifferent([A1,A2,B1,B2])*, is equivalent to defining the constraints in the pseudo code manner below:

$$A1 \neq A2, A1 \neq B1, A1 \neq B2,$$
$$A2 \neq B1, A2 \neq B2, B1 \neq B2$$

**ssRowsAllDifferent($C_{TL}$:$C_{BR}$)** : The specific $C_{TL}$:$C_{BR}$ cell range syntax is the only cell range reference that is allowed with this function. It means that the cell variables in each row will have different values. In the example below, the table represents the setting of the domain variables values and the example will show the *ssRowsAllDifferent* function with the matrix cell range.

|   | A | B |
|---|---|---|
| 1 | **1..6** | **1..4** |
| 2 | **1..4** | **1..5** |
| 3 | **1..4** | **1..5** |

*Example 1*:

The *ssRowsAllDifferent(A1:B3)*, is equivalent to defining the constraints in the pseudo code manner below:

For row 1: A1 ≠ B1
For row 2: A2 ≠ B2
For row 3: A3 ≠ B3

**ssColsAllDifferent($C_{TL}$:$C_{BR}$)**   : The specific $C_{TL}$:$C_{BR}$ cell range syntax is the only cell and range reference that is allowed with this function. It means that the cell variables in each column will have different values

|   | A | B |
|---|---|---|
| 1 | **1..6** | **1..4** |
| 2 | **1..4** | **1..5** |
| 3 | **1..4** | **1..5** |

*Example 1*:

The *ssColsAllDifferent(A1:B3)*, is equivalent to defining the constraints in the pseudo code manner below:

For column A:  A1 ≠ A2, A1 ≠ A3, A2 ≠ A3
For column B:  B1 ≠ B2, B1 ≠ B3, B2 ≠ B3

**ssColsAggregate($BinArithOp$,$C_{TL}$:$C_{BR}$, $RelOp$,$ResultList$):**  this spreadsheet-specific function works in the following manner:

1.  Aggregate the columns in $C_{TL}$:$C_{BR}$ according to the binary arithmetic operator ***BinArithOp***.

2.  Then evaluate each aggregated column in $C_{TL}$:$C_{BR}$, to its equivalent *item* in the ***ResultList*** depending on the relational operator ***RelOp***.

3.  The rules of interaction are as follows:

    *   If the number of *items* in the ***ResultList*** is less than the number of columns in the cell matrix $C_{TL}$:$C_{BR}$, then the last *item* in the ***ResultList*** will be used to satisfy the remaining columns.

    *   If the number of *items* in the ***ResultList*** is greater than the number of columns in the cell matrix $C_{TL}$:$C_{BR}$, then excess *items* in the ***ResultList*** will not be used.

    *   If all the *items* in the ***ResultList*** are the same value, then just one *item* can be entered.

The table below with the domain variables and values will be used in the examples below:

|   | A | B | C | D | E |
|---|---|---|---|---|---|
| 1 | **1** | **2** | **3** | **1** | **2** |
| 2 | **1** | **1** | **1** | **1** | **1** |
| 3 | **2** | **5** | **3** | **2** | **1** |
| 4 | **1** | **1** | **1** | **1** | **1** |

*Example 1*:

The *ssColsAggregate(+,A1:E2, #=,[1,0,1,1,2])*, is equivalent to defining the constraints in the pseudo code manner below, since there are five columns in cell range matrix A1:E2, the ***ResultList*** will have five corresponding values with the ***BinArithOp*** symbol being a + and the ***RelOp*** being a #= symbol.

$$A1 + A2 \; \#= 1$$
$$B1 + B2 \; \#= 0$$
$$C1 + C2 \; \#= 1$$
$$D1 + D2 \; \#= 1$$
$$E1 + E2 \; \#=2$$

*Example 2*:

The *ssColsAggregate(+,A1:E2, #>,1)*, is equivalent to defining the constraints in the pseudo code manner below. Since **ResultList** have all the same values, only one iteration of value is required.

$$A1 + A2 \; \#> 1$$
$$B1 + B2 \; \#> 1$$
$$C1 + C2 \; \#> 1$$
$$D1 + D2 \; \#> 1$$
$$E1 + E2 \; \#> 1$$

*Example 3*:

The *ssColsAggregate(+,A1:B2, #>,C1:C3)*, is equivalent to defining the constraints in the pseudo code manner below. Since **ResultList** will have 3 columns, which would be more than the number of columns in the cell range matrix A1:B2 which is 2, therefore only 2 fo the column values in **ResultList** will be used.

$$A1 + A2 \; \#> 3$$
$$B1 + B2 \; \#> 1$$

**ssRowsAggregate($BinArithOp$,$C_{TL}$:$C_{BR}$, $RelOp$,$ResultList$):** this spreadsheet-specific function works in a similar manner to the *ssColsAggregate* function:

1. Aggregate the rows in $C_{TL}$:$C_{BR}$ according to the binary arithmetic operator **BinArithOp**.

2. Then evaluate each aggregated rows in $C_{TL}$:$C_{BR}$, to its equivalent item in the **ResultList** depending on the relational operator **RelOp**.

3. The rules of interaction are as follows:

- If the number of *items* in the **ResultList** is less than the number of rows in the cell matrix $C_{TL}$:$C_{BR}$, then the last *item* in the **ResultList** will be used to satisfy the remaining rows.

- If the number of *items* in the **ResultList** is greater than the number of rows in the cell matrix $C_{TL}$:$C_{BR}$, then excess *items* in the **ResultList** will not be used.

- If all the *items* in the **ResultList** are the same value, then just one *item* can be entered.

The table below with the domain variables and values will be used in the examples below:

|   | A | B | C | D | E |
|---|---|---|---|---|---|
| 1 | **1** | **2** | **3** | **1** | **2** |
| 2 | **1** | **1** | **1** | **1** | **1** |
| 3 | **2** | **2** | **3** | **7** | **9** |
| 4 | **1** | **1** | **1** | **1** | **1** |

*Example 1*:

The *ssRowsAggregate(+,A1:E2, #\=,[1,3])*, is equivalent to defining the constraints in the pseudo code manner below, since there are two rows in the cell range matrix A1:E2, the **ResultList** will have 2 corresponding values with the **BinArithOp** symbol being a + and the **RelOp** being a #\= symbol.

$$A1 + B1 + C1 + D1 + E1 \; \#\backslash= 1$$
$$A2 + B2 + C2 + D2 + E2 \; \#\backslash= 3$$

*Example 2*:

The *ssRowsAggregate(+,A1:E2, #=,1)*, is equivalent to defining the constraints in the pseudo code manner below, since there are two rows in A1:E2, the **ResultList** will have 1 corresponding values with the **BinArithOp** symbol being a + and the **RelOp** being a #= symbol. Since **ResultList's** values are all the same, only one iteration of it is required to be stated.

$$A1 + B1 + C1 + D1 + E1 \#= 1$$
$$A2 + B2 + C2 + D2 + E2 \#= 1$$

*Example 3*:

The *ssRowsAggregate(+,A1:B2, #\=,D1)*, is equivalent to defining the constraints in the pseudo code manner below, since there are two rows in the matrix A1:E2, the **ResultList** has only 1 corresponding value with the **BinArithOp** symbol being a + and the **RelOp** being a #\= symbol.

$$A1 + B1 + C1 \ \#\= 1$$
$$A2 + B2 + C2 \ \#\= 1$$

**ssPairCellsAggregate($C_{TL1}$:$C_{BR1}$, BinArithOp, $C_{TL2}$:$C_{BR2}$, RelOp,ResultList):** This function is used to aggregate two disparate cell matrix ranges noted by $C_{TL1}$:$C_{BR1}$ and $C_{TL2}$:$C_{BR2}$ in the following manner:

1. Aggregate the disparate cells in $C_{TL1}$:$C_{BR1}$ and $C_{TL2}$:$C_{BR2}$ according to the binary arithmetic operator **BinArithOp**.

2. Then evaluate each newly paired cell to its equivalent *item* in the **ResultList** depending on the relational operator **RelOp**.

3. The rules of interaction are as follows:

   - If the number of *items* in the **ResultList** is less than the number of newly paired cells, then the last *item* in the **ResultList** will be used to satisfy the remaining paired cells.

- If the number of *items* in the **ResultList** is greater than the number of newly paired cells, then the excess *items* in the **ResultList** will just be ignored.

- If all the items in the **ResultList** are the same value, then just one item can be entered.

The table below with the domain variables and values will be used in the examples below:

|   | A | B | C | D | E |
|---|---|---|---|---|---|
| 1 | 1 | 2 | 3 | 1 | 2 |
| 2 | 1 | 1 | 1 | 1 | 1 |
| 3 | 1 | 2 | 4 | 8 | 1 |
| 4 | 2 | 2 | 1 | 3 | 3 |
| 5 | 1 | 6 | 8 | 9 | 0 |
| 6 | 1 | 1 | 1 | 1 | 1 |
| 7 | 1 | 1 | 1 | 1 | 1 |

*Example 1*:

*ssPairCellsAggregate(A1:B3, +, C5:D7, #>, [1,2,3,4,6,1])* will evaluate to the constraints below:

$$A1 + C5 \text{ \#> } 1,$$
$$B1 + D5 \text{ \#> } 2$$
$$A2 + C6 \text{ \#> } 3,$$
$$B2 + D6 \text{ \#> } 4$$
$$A3 + C7 \text{ \#> } 6,$$
$$B3 + D7 \text{ \#> } 1$$

*Example 2*:

*ssPairCellsAggregate(A1:B3, +, C5:D7, #\=, 1)* will evaluate to the constraints below:

$$A1 + C5 \text{ \#\= } 1,$$
$$B1 + D5 \text{ \#\= } 1$$
$$A2 + C6 \text{ \#\= } 1,$$
$$B2 + D6 \text{ \#\= } 1$$
$$A3 + C7 \text{ \#\= } 1,$$

$$B3 + D7 \; \#\backslash= 1$$

*Example 3*:

*ssPairCellsAggregate(A1:B3,* +**,** C5:D7**,** #>**,** D3:E4) will evaluate to the constraints six

number of pair cells derived from the matrices A1:B3 and C5:D7, and the **ResultList**

D3:E4 will have only four, so the last value will be repeated two more times:

$$A1 + C5 \; \#> 8,$$
$$B1 + D5 \; \#> 1,$$
$$A2 + C6 \; \#> 3,$$
$$B2 + D6 \; \#> 3$$
$$A3 + C7 \; \#> 3,$$
$$B3 + D7 \; \#> 3$$

**ssDiagonalAggregate(*BinArithOp*, $C_{TL}$:$C_{BR}$, *RelOp*,*ResultList*):** This function is used to

aggregate diagonal cells in cell range matrix $C_{TL}$:$C_{BR}$ in the following manner:

1. Aggregate the diagonal cells in $C_{TL}$:$C_{BR}$ according to the binary arithmetic

   operator *BinArithOp*.

2. Then evaluate each aggregated diagonal cell to its equivalent *item* in the

   **ResultList**, depending on the relational operator *RelOp*.

3. The rules of interaction are as follows:

   - If the number of *items* in the **ResultList** is less than the number of newly

     aggregated diagonal cells, then the last *item* in the **ResultList** will be used

     to satisfy the remaining diagonal cells.

   - If the number of *items* in the **ResultList** is greater than the number of

     aggregated diagonal cells, then the excess *items* in the **ResultList** will just

     be ignored.

- If all the items in the **ResultList** are the same value, then just one item can be entered.

The table below with the domain variables and values will be used in the examples below:

|   | A | B | C | D | E |
|---|---|---|---|---|---|
| 1 | **1** | **2** | **3** | **1** | **2** |
| 2 | **0** | **2** | **0** | **0** | **0** |
| 3 | **0** | **2** | **3** | **0** | **0** |
| 4 | **1** | **1** | **1** | **4** | **1** |

*Example 1*:

*ssDiagonalAggregate(+,A1:E4 , #>,* [1,2,3,1,1,5,1,1]) will evaluate to the constraints below:

E1 #> 1,
D1 + E2 #> 2,
C1 + D2 + E3 #> 3
B1 + C2 + D3 + E4 #> 1,
A1 + B2 + C3 + D4 #> 1
A2 + B3 + C4 #> 5,
A3 + B4 #> 1,
A4 #> 1

*Example 2*:

*ssDiagonalAggregate(+,A1:E4 , #>,* 1) will evaluate to the constraints below:

E1 #> 1,
D1 + E2 #> 1,
C1 + D2 + E3 #> 1
B1 + C2 + D3 + E4 #> 1,
A1 + B2 + C3 + D4 #> 1
A2 + B3 + C4 #> 1,
A3 + B4 #> 1,
A4 #> 1

**ssBackDiagonalAggregate(*BinArithOp*, *C_{TL}*:*C_{BR}*, *RelOp*,*ResultList*):** This function is used to aggregate diagonal cells in cell matrix $C_{TL}$:$C_{BR}$ in the following manner:

1. Aggregate the diagonal cells in $C_{BR}$:$C_{TL}$ according to the binary arithmetic operator ***BinArithOp***.

2. Then evaluate each aggregated diagonal cell to its equivalent *item* in the ***ResultList***, depending on the relational operator ***RelOp***.

3. The rules of interaction are as follows:

   - If the number of *items* in the ***ResultList*** is less than the number of newly aggregated diagonal cells, then the last *item* in the ***ResultList*** will be used to satisfy the remaining diagonal cells.

   - If the number of *items* in the ***ResultList*** is greater than the number of aggregated diagonal cells, then the excess *items* in the ***ResultList*** will just be ignored.

   - If all the items in the ***ResultList*** are the same value, then just one item can be entered.

The table below with the domain variables and values will be used in the examples below:

|   | A | B | C | D | E |
|---|---|---|---|---|---|
| 1 | 1 | 2 | 3 | 1 | 2 |
| 2 | 0 | 2 | 0 | 0 | 0 |
| 3 | 0 | 2 | 3 | 0 | 0 |
| 4 | 1 | 1 | 1 | 4 | 1 |

*Example 1:*

*ssBackDiagonalAggregate(+,A1:E4 , #>,* [1,2,3,1,1,5,1,1]) will evaluate to the constraints below:

```
A1 #> 1,
B1 + A2 #> 2,
C1 + B2 + A3 #> 3
D1 + C2 + B3 + A4 #> 1
E1 + D2 + C3 + B4 #> 1
E2 + D3 + C4 #> 5
E3 + D4 #> 1
E4 #> 1
```

**ssMin(*CellRange*)** : This spreadsheet-specific function is used to state that the cell value in ***CellRange***, which has to be a single value should be the smallest possible value for the constraint problem. By setting the constraint *ssMin(CellRange)*, the cell variables will be labeled in such a way that ***CellRange*** assumes the smallest possible value.

**ssMax(*CellRange*)** : This spreadsheet-specific function is similar to the *ssMin(**CellRange**)* function, with the only difference being that it assumes the cell variable in ***CellRange*** to be the maximum possible value.

### 4.2.2.3 Framework required spreadsheet specific Functions

There are spreadsheet-specific functions that are needed to get the cell ranges of the domains and constraints, so that the parser could easily build the CLP program. The absence of these functions will require the parsing of the entire spreadsheet, and additional logic to separate the domains and constraints and put them together to form a coherent CLP program. The two specific functions with their overloaded combinations are called *ssVarRanges*() used to specify variable ranges and *ssConstraintRanges*() used to specify constraint variable ranges .

***ssVarRanges(CellRange):*** This is used to specify that the variables are located in the

***CellRange*** reference. The examples below show the various combinations.

<u>*Example 1:*</u>

*ssVarRanges(A1)* : specifies that the domain variable is located in the cell with a column

A and row1.

<u>*Example 2:*</u>

*ssVarRanges(A1:B2)* : specifies that the domain variables are located in the range of cells in columns A through B and rows 1 through 2.

<u>*Example 3:*</u>

*ssVarRanges(A1,B2,C1)* : specifies that the domain variables are located enumerated cells

A1, B2, and C1.

<u>*Example 4:*</u>

*ssVarRanges(A1:B2,Sheet2!C1:D2)* : specifies that the domain variables are located in

the current worksheet range A1:B2, and also on worksheet2 named Sheet2's range

C1:D2.

**ssConstraintRanges** (***CellRange***) **:** This is used to specify that the constraints are located in the ***CellRange*** reference. The examples below show the various combinations.

<u>*Example 1:*</u>

*ssConstraintRanges(A1)* : specifies that the constraint is located in the cell with a column

A and row1.

<u>*Example 2:*</u>

*ssConstraintRanges(F1:G2)* : specifies that the constraints are located in the range of cells in columns F through G and rows 1 through 1.

|   | F | G |
|---|---|---|
| 1 | **ssDomain(A1:B2,1,3)** | |
| 2 | | **ssAllDifferent(A1:B2)** |

*Example 3:*

*ssConstraintRanges(A1:B2,Sheet2!C1:D2)* : specifies that the constraints are located in the current worksheet range A1:B2, and also on worksheet2 named Sheet2's range C1:D2.

These functions will allow users to separate the domain variable and constraint declaration into smaller more manageable sections, further aiding the handling of scalability and complexity.

## 4.3 Language Communication Layer

The Language Communication layer exists between the *Spreadsheet layer* and the *Constraint Solver layer*.  As shown in Fig 4.6, it handles CHR/Excel library additions, and makes sure that the parser/analyzer gets the required information from both the CHR/Excel library and Transformation Rules.

The Parser/Analyzer analyzes the gathered information and parses it into a correctly formed program that the constraint solver can compile. It is also responsible for running the given parameters, and returning the results to the spreadsheet interface through the Query Result Interpreter.  Since we are dealing with the SWI as the constraint solver, the syntax that will be used will be close to SWI Prolog.

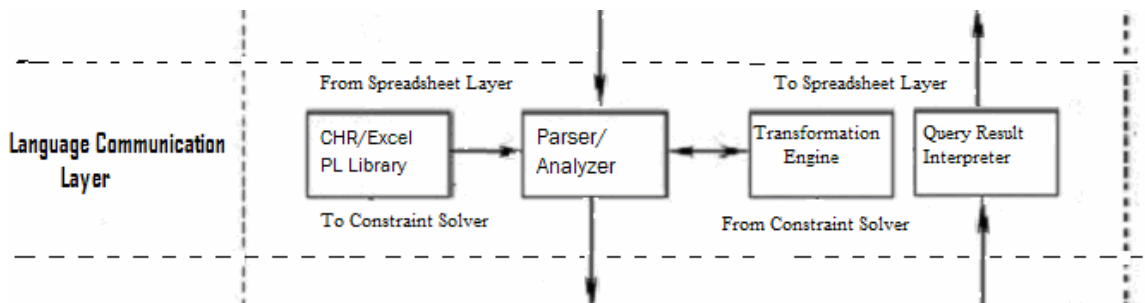

Fig 4.6 Language Communication Layer design

### 4.3.1 Parser/Analyzer

The Parser/Analyzer gathers all the necessary input information namely: spreadsheet values, CHR/Excel PL and Language compiler information, and builds the appropriate constraint logic programming language file with a .pl extension. It parses through the Excel's A1 reference style [18] information, builds domain variable information, and converts the spreadsheet specific functions into SWI-Prolog's syntax. It then writes to a file with the primary/current worksheet's name and a .pl extension as illustrated in Fig 4.7.

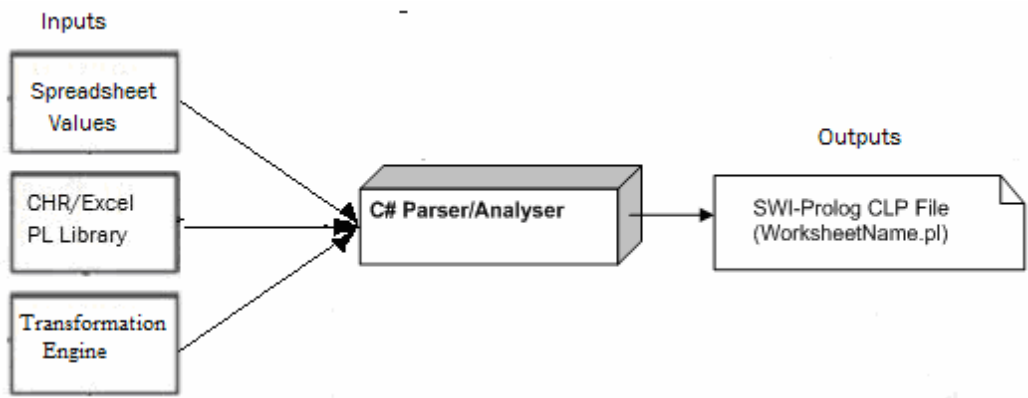

Fig 4.7 Illustration of .pl file creation

The parser/analyzer is also responsible for maintaining the states for the original spreadsheet state before the parsing began. This allows users to quickly go back to the original state, incase they want to change the problem model in order to derive the correct one or new solution. In addition, the parser/analyzer also maintains all the derived solutions to allow users to dynamically view or iterate through all of them, until the desired one is found.

## 4.3.2 Transformation Engine

The transformation engine consists of rules that are going to be divided into three main sub-divisions. The first covers the cell range transformation into a complete list of variables, the second will covers the Domain transformation and the third will cover the spreadsheet-specific constraint transformation, using components from the first sections.

### 4.3.2.1 CellRange Transformation

This sub section covers algorithms that will take spreadsheet variables and convert them into lists and sub-lists. The first algorithm covers the creation of a list of variables. The second covers the creation of a list containing sub-lists grouped by rows, the next algorithm is similar to the second with the difference being that the sub-lists are grouped by columns. The last two algorithms cover the creation of a list containing sub-lists group by diagonal or back diagonal cell variables.

 Algorithm to create a list of variables

The algorithm in Fig 4.8 takes spreadsheet variables as input and creates a list of SWI-Prolog cell variables output.  The *extractRowId* (C) extracts the row id (e.g. for cell A1 the *extractRowId* (A1) will return A), and *extractColId* (A1) will return 1.

LIST-CREATION
1.  INPUT: C          % takes any of the styles defined in the CellRange definition section 4.2.2
2.  **if** *IsMatrix*(C) **then**
3.      $L \leftarrow [\ ]$
4.      **for** $r \leftarrow extractRowId(C_{TL})$ to $extractRowId(C_{BR})$
5.          **for** $c \leftarrow extractColId(C_{TL})$ **to** $extractColId(C_{BR})$
6.              $L \leftarrow add(cr, L)$
7.  **else if** *IsList*(C) **then**
8.          $L \leftarrow C$
9.      **else**
10.         $L \leftarrow [C]$
11. OUTPUT: $L$

---

Fig 4.8 List Creation algorithm

The function proceeds as follows:

- It begins by receiving the input **CellRange** C.

- It then checks to see if the parameter **CellRange** is defined as a matrix, which are the formats defined in the *table below*.

| Input | Output |
|---|---|
| A1:B2 | [A1, B1, A2, B2] |
| Sheet2!A1:B2 | [Sheet2A1, Sheet2B1, Sheet2A2, Sheet2B2] |

- If the **CellRange** is in a matrix format (Line 2), it will extract the row-id of both the top left and bottom right cells, and then produce the output shown in the table above using two nested for loops. E.g. **CellRange** A1:B2 is equivalent to A1 being $C_{TL}$ and B2 equal to $C_{BR}$. The top **for** loop will be from 1 to 2 and the inner **for** loop from A to B, and L1 will become a list of variables shown in Line 6.

- If the **CellRange** variable is a single variable as shown in the *table below*'s A1, it will add it to the list and output it. If it is already a list, it will just return that list as shown in the output for the second row below.

| Input | Output |
|---|---|
| A1 | [A1] |

| [A1, A5, B6] | [A1, A5, B6] |
|---|---|

Algorithm to create a list of variables grouped by row

The algorithm in Fig 4.9 takes spreadsheet variables as input and creates a SWI-Prolog equivalent list of variables with *cells in the same row* being placed into sub-lists.

---

ROW-LIST-CREATION
1. INPUT: C  % ONLY takes a matrix CellRange parameter, all others will return an error message
2. **if** *IsMatrix*(C) **then**
3.   $L1 \leftarrow [\ ]$
4.   **for**  $r \leftarrow$ *extractRowId*($C_{TL}$) **to** *extractRowId*($C_{BR}$)
5.     $L2 \leftarrow [\ ]$
6.     **for**  $c \leftarrow$ *extractColId*($C_{TL}$) **to** *extractColId*($C_{BR}$)
7.       $L2 \leftarrow add(cr, L2)$
8.     $L1 \leftarrow add([L2], L1)$
9.     OUTPUT: *L1*
10. **else**
11.   OUTPUT: *Show-Error-Msg()*

---

Fig 4.9 Row List Creation algorithm

The function proceeds as follows:

- This function is similar to the LIST-CREATION function, with the algorithm

  containing another list variable name *L2* shown in **Line 5**.

- If we look at the pseudocode from **Line 6-7**, *L2* responsible for placing cells in

  the same row in its list.

- **Line 8** takes the *L2* sub-list and adds it into the main list *L1*. The table below

  shows the input consisting of the spreadsheet matrix ***CellRange*** and its equivalent

  output.

| Input | Output |
|---|---|
| A1:B3 | [ [A1,B1],[A2,B2],[A3,B3]] |
| Sheet2!A1:B2 | [[Sheet2A1, Sheet2B1], [Sheet2A2, Sheet2B2]] |

Column List Creation Algorithm

The algorithm in Fig 4.10 takes spreadsheet variables as input and creates a SWI-Prolog equivalent list with *cells in the same column* being placed into sub-lists.

---

COLUMN-LIST-CREATION
1. INPUT: C          % ONLY takes a matrix CellRange parameter; all others will return an error message
2. **if** *IsMatrix*(C) **then**
3.   *L1* ← [ ]
4.   **for**  *c* ← *extractColId*(C$_{TL}$) **to**  *extractColId*(C$_{BR}$)
5.   *L2* ← [ ]
6.     **for**  *r* ← *extractRowId*(C$_{TL}$) **to**  *extractRowId*(C$_{BR}$)
7.               *L2* ← *add*(*L2, cr*)
8.   *L1* ← *add*([*L2*],*L1*)
9.   OUTPUT: *L1*
10. **else**
11.   OUTPUT: *Show-Error-Msg()*

---

Fig 4.10 Column List Creation algorithm

The function proceeds as follows:

- This function is similar to the ROW-LIST-CREATION function, with the

  algorithm responsible for placing cells in the column in a sub-list and

  concatenating it to the main list. The table below shows the input and output for

  this function.

| Input | Output |
|---|---|
| A1:B3 | [[A1,A2,A3],[B1,B2,B3]] |
| Sheet2!A1:B2 | [[Sheet2A1, Sheet2B1], [Sheet2A2, Sheet2B2]] |

Diagonal List Creation Algorithm

The algorithm in Fig 4.11 takes spreadsheet variables as input and creates a SWI-Prolog equivalent list with diagonal *cells starting from the left hand side* being placed into sub-lists.

---

DIAGONAL-LIST-CREATION

1. INPUT: C  % ONLY takes a matrix CellRange parameter, all others will return an error message
2. **if not(** *IsMatrix*(C)) **then**
3.   OUTPUT: *Show-Error-Msg()*
4. %% C is in a Matrix form as $C_{TL}$:$C_{BR}$
5. $L$ = [ ]
6. **for** $c \leftarrow$ *extractColId*($C_{TL}$) **to** *extractColId*($C_{BR}$)
7.   *L1* ← [ ]
8.   $c_i$ ← $c$
9.   $r$ ← *extractRowId*($C_{TL}$)
10.   **while** ( $r \leq$ *extractRowId($C_{BR}$)* **and** $c_i \leq$ *extractColID($C_{BR}$)* )
11.     *L1* ← *add($c_i r$, L1)*
12.     $c_i$ ++
13.     $r$ ++
14.   $L$ ← *add(L1, L)*
15.
16. **for** $r \leftarrow$ *extractRowId*($C_{TL}$) + 1 **to** *extractRowId*($C_{BR}$)
17.   *L1* ← [ ]
18.   $r_i$ ← $r$
19.   $c$ ← *extractColId*($C_{TL}$)
20.   **while** ( $r_i \leq$ *extractRowId($C_{BR}$)* **and** $c \leq$ *extractColId($C_{BR}$)* )
21.     *L1* ← *add($cr_i$, L1)*
22.     $r_i$ ++
23.     $c$ ++
24.   $L$ ← *add(L1, L)*
25. OUTPUT: $L$

---

Fig 4.11 Diagonal List Creation algorithm

The function proceeds as follows:

- It begins by receiving the input ***CellRange*** C.

- It then checks to see if the parameter ***CellRange*** is defined as a matrix. If it isn't it will output an error message as shown in **Line 3**.

- The first loop that begins on **Line 6** and ends on **Line 14** is used to calculate the top triangular half of the matrix. The table below will have the top triangular half matrix marked in a bold font and will generate the following sub-list assigned to *L1* and add it to the main list *L*:

L ← [[A1, B2, C3, D4, E5], [B1, C2, D3, E4], [C1, D2, E3], [D1, E2], [E1]]

|   | A | B | C | D | E |
|---|---|---|---|---|---|
| 1 | **1** | **7** | **3** | **3** | **4** |
| 2 | 2 | **7** | **4** | **4** | **7** |
| 3 | 2 | 3 | **4** | **3** | **6** |
| 4 | 3 | 5 | 4 | **2** | **6** |
| 5 | 6 | 6 | 8 | 8 | **7** |

- **Line 16** through **Line 23,** will loop through and generate the bottom triangular half sub-list L1. The generated sub-lists are shown below:

  [A2, B3, C4, D5], [A3, B4, C5], [A4, B5], [A5]

- **Line 24** has each sub-list generated added to L, as each sub-list is created with a final list L with diagonal sub-lists shown below:

  L ← [[A1, B2, C3, D4, E5], [B1, C2, D3, E4], [C1, D2, E3], [D1, E2], [E1],

  [A2, B3, C4, D5], [A3, B4, C5], [A4, B5], [A5]]

Back-Diagonal List Creation Algorithm

The algorithm in Fig 4.12 takes spreadsheet variables as input and creates a SWI-Prolog equivalent list with diagonal *cells starting from the right hand side* being placed into sub-lists.

---

BACK-DIAGONAL-LIST-CREATION
1. INPUT: C        % ONLY takes a matrix CellRange parameter; all others will return an error message
2. **if  not(** *IsMatrix*(C)) **then**
3.      OUTPUT: *Show-Error-Msg()*
4. %% C is in a Matrix form as $C_{TL}$:$C_{BR}$
5. *L* = [ ]
6. **for**  $c \leftarrow$ *extractColId*($C_{BR}$) − 1 to  *extractColId*($C_{TL}$)

7.      $L1 \leftarrow [\,]$

8.      $c_1 \leftarrow c$

9.      $r \leftarrow extractRowId(C_{TL})$

10.     **while** ( $r < extractRowId(C_{BR})$ **and** $c_1 \geq extractColId(C_{TL})$ )

11.          $L1 \leftarrow add(c_1r, L1)$

12.          $c_1$ --

13.          $r$ ++

14.    $L \leftarrow add(L1, L)$

15.

16. **for** $r \leftarrow extractRowId(C_{TL})$ **to** $extractRowId(C_{BR})$

17.    $L1 \leftarrow [\,]$

18.    $r_1 \leftarrow r$

19.    $c \leftarrow extractColId(C_{BR})$

20.    **while** ( $r_1 \leq extractRowId(C_{BR})$ **and** $c \geq extractColId(C_{TL})$ )

21.         $L1 \leftarrow add(cr_1, L1)$

22.         $r_1$ ++

23.         $c$ --

24.    $L \leftarrow add(L1, L)$

25. OUTPUT: $L$

---

Fig 4.12 Back-Diagonal List Creation algorithm

The function proceeds as follows:

- It is similar to the DIAGONAL-LIST-CREATION, with the difference being the loop in **Line 6**, evaluating from the column of the *CellRange* matrix's bottom right hand side to its top left side. This allows it to get the bottom triangular half starting from the right hand side.

- The second loop from **Line16** to **Line 23** is used to calculate the top triangular half of the matrix, starting from the top left side. The table below shows the top left triangular side marked in bold:

|   | A | B | C | D | E |
|---|---|---|---|---|---|
| 1 | **1** | **7** | **3** | **3** | 4 |
| 2 | **2** | **7** | **4** | 4 | 7 |
| 3 | **2** | **3** | 4 | 3 | 6 |
| 4 | **3** | 5 | 4 | 2 | 6 |
| 5 | 6 | 6 | 8 | 8 | 7 |

- **Line 24** has each sub-list generated added to L, as each sub-list is created with a final list L with diagonal sub-lists shown below:

  L ← [[E1, D2, C3, B4, A5] [E2, D3, C4, B5], [E3, D4, C5], [E4, D5], [E5],

  [D1, C2, B3, A4], [C1, B2, A3], [B1, A2], [A1]]

## 4.3.2.2 Domain Transformation

The algorithm below transforms the domain variable cell values to the SWI-Prolog equivalent, with the input variable and integer cell value producing the SWI-Prolog equivalent as shown in the table below:

---

DOMAIN-TRANSFORMATION
1.  INPUT1:  C      % domain cell variable
2.  INPUT2:  V     % integer cell value
4.  **if** *SingleIntegerValue*(V) **then**
5.      OUTPUT : C #= V
6.  **else**
7.      OUTPUT : C in V

---

Fig 4.13 Domain Transformation algorithm

| Variable | Set of Integer Cell Value | SWI-Prolog equivalent |
|----------|---------------------------|------------------------|
| A1 | 200 | A1 #= 200 |
| A1 | 1..3 | A1 in 1..3 |
| A1 | [1,2,3,5,6] | A1 in [1,2,3,5,6] |

## 4.3.2.3 Spreadsheet-Specific Constraint Transformations

This section describes the algorithms that will be used to transform the spreadsheet-specific constraint functions into their SWI-Prolog equivalent.

Algorithm on Domain Constraints

The algorithm in Fig 4.14 takes spreadsheet-specific *ssDomain* function and creates its

SWI-Prolog equivalent.

---

SS-DOMAIN-TO-PROLOG
1. INPUT: *ssDomain(C,Min,Max)*
3. $L \leftarrow$ LIST-CREATION(C)
4. OUTPUT: *L in Min..Max*

---

Fig 4.14 ssDomain to Prolog Conversion

The function then proceeds as follows:

- Takes the *ssDomain(C,Min,Max)* spreadsheet-specific function as input, then

  sends the **CellRange** matrix *C* to the LIST-CREATION function and returns a list

  of cell variables *L*. It then outputs the list *L* with the min, max values

  concatenated. The table below shows the input and output values of the algorithm.

| Input | Output |
|---|---|
| ssDomain([A1,A2,B3],1,5) | [A1,A2,B3] in 1..5 |
| ssDomain(A1:B2,1,3) | [A1,B1,A2,B2] in 1..5 |

Algorithm to convert ssAllDifferent to Prolog equivalent

The algorithm in Fig 4.15 takes spreadsheet-specific function *ssAllDifferent* and creates its SWI-Prolog function equivalent named *all_different()*.

---

SS-ALL-DIFFERENT-TO-PROLOG(C)
1. INPUT: *ssAllDifferent(C)*
2. $L \leftarrow$ LIST-CREATION(C)
3. OUTPUT: *all_different(L)*

---

Fig 4.15 ssAllDifferent to Prolog Conversion

The function then proceeds as follows:

- Takes the *ssAllDifferent(C)* spreadsheet-specific function as input, then sends the **CellRange** matrix *C* to the LIST-CREATION function and returns a list of cell variables *L*. It then outputs the prolog equivalent *all_different(L)*. The table below shows the input and output values of the algorithm.

| Input | Output |
|---|---|
| ssAllDifferent([A1,B2,E5]) | all_different([A1,B2,E5]) |
| ssAllDifferent(A1:B2) | all_different([A1,B1,A2,B2]) |

Algorithm to convert ssColsAllDifferent to Prolog equivalent

The algorithm in Fig 4.16 takes spreadsheet-specific *ssColsAllDifferent* function and creates its SWI-Prolog equivalent.

---

SS-COLS-ALL-DIFF-TO-PROLOG(C)
1. INPUT: *ssColsAllDifferent(C)*
2. *L* ← COLUMN-LIST-CREATION(C)
3. OUTPUT: *subListAllDifferent(L)*

---

Fig 4.16 ssColsAllDifferent to Prolog Conversion

The function then proceeds as follows:

- This function takes the spreadsheet-specific function *ssColsAllDifferent(C)*. It then sends the **CellRange** matrix *C* as input to the COLUMN-LIST-CREATION, which returns a list of cell variable values *L*. It then proceeds to call the Prolog function *sublistAllDifferent* which recursively will create the all_different() function. The algorithm in Fig 4.17 shows the syntax of the *sublistAllDifferent* function.

**Prolog Predicate located in the Excel PL library**

```
sublistAllDifferent([]).
sublistAllDifferent([F| R]) :-  all_different(F),
                                sublistAllDifferent(R).
```

Fig 4.17 subListAllDifferent Prolog Predicate.

- The table below shows this functions input and output values.

| Input | Prolog-Predicate Output | Output |
|-------|------------------------|--------|
| ssColsAllDifferent(A1:B3) | sublistAllDifferent([[A1,A2,A3], [B1,B2,B3] ]) | all_different([A1,A2,A3]) all_different([B1,B2,B3]) |

Algorithm to convert ssRowsAllDifferent to Prolog equivalent

The algorithm in Fig 4.18 takes spreadsheet-specific *ssRowsAllDifferent* function and creates its SWI-Prolog equivalent.

SS-ROWS-ALL-DIFF-TO-PROLOG(C)
1. INPUT: *ssRowsAllDifferent(C)*
2.   *L* ← ROW-LIST-CREATION(C)
3. OUTPUT: *subListAllDifferent(L)*

Fig 4.18 ssRowsAllDifferent to Prolog Conversion

The function then proceeds as follows:

- This function is similar to the SS-ALL-DIFFERENT-TO-PROLOG function, with the difference being that the ROW-LIST-CREATION function is used instead of the PROLOG-LIST-CREATION function.

- The table below shows this functions input and output values.

| Input | Output |
|-------|--------|
| ssRowsAllDifferent(A1:B3) | all_different([A1,B1]) all_different([A2,B2]) all_different([A3,B3]) |

<u>Algorithm to convert ssColsAggregate to Prolog equivalent</u>

The algorithm in Fig 4.19 takes spreadsheet-specific *ssColsAggregate* function and creates its SWI-Prolog equivalent.

---

SS-COLS-AGGREGATE-TO-PROLOG
1. INPUT: *ssColsAggregate (B_0,C_{TL}:C_{BR},R_0,L)*
2. *L1 ←* COLUMN-LIST-CREATION(*C_{TL}:C_{BR}*)
3. *N ← length(L1)*
4. *L2 ←* SETLEN(*L, N*)
5. OUTPUT: *subListAggregate(B_0, L1, R_0, L2)*

---

Fig 4.19 ssColsAggregate to Prolog Conversion

The function then proceeds as follows:

- This function takes the spreadsheet-specific function *ssColsAggregate*. It then

  sends the **CellRange** matrix $C_{TL}:C_{BR}$ as input to the COLUMN-LIST-

  CREATION, which returns a list of cell variable values *L1*. It then proceeds to get

  the length of the list, which is passed to the *SETLEN* function whose algorithm is

  shown in Fig 4.20. The result of the *SETLEN* function, is then used as a parameter

  of the *subListAggregate* Prolog function which shown in Fig 4.21.

---

SETLEN
1. INPUT: *L* , N
2. *L1 ←* [ ]
3. **if** *length(L) > N* **then**
4.     **for** i ← 1  **to**  *N*
5.         *L1 ← add(L[i], L1)*
6. **else**
7.     *N1 ← N – length(L)*
8.     *L1 ← L*
9.     *Item ← L[length(L)]*

10.     **for**  *i* ←1 **to** *N1*
11.      *L1 ← add(item, L1)*
12.  OUTPUT  *L1*

---

Fig 4.20 *SETLEN* function algorithm

- The SETLEN function is used to expand the contents of the list *L* according the *length N*. If the list *L* is larger than the *N*, then a new list *L1* is created containing the elements of *L* up to *N*. Otherwise a new list *L1* is created containing all the elements of *L*, with the last element of L, being added *N1* (which is a difference between N and the length of *L*) times to the new list *L1*.

| Prolog Predicate located in the Excel PL library |
|---|
| subListAggregate(B0,[],_,_).<br>subListAggregate(B0,[F1\| R1], R0, [F2 \| R2]) :-<br>                  aggregate(B0,F1, R0, F2),<br>                  subListAggregate(B0,R1, R0,R2). |

Fig 4.21 *subListAggregate Prolog predicate*

- The *subListAggregate* prolog predicate is used to recursively format the ssColsAggregate spreadsheet-specific function into the appropriate prolog syntax, with the help of another prolog predicate named aggregate whose algorithm is shown in Fig 4.22.

| Prolog Predicate located in the Excel PL library |
|---|
| aggregate(Bo, [F \| T], Ro, V) :-<br>                aggrConstraint(Bo, T, F, S),<br>                S1 =.. [Ro, S, V],<br>                S1.<br><br>aggrConstraint(_, [], S, S).<br>aggrConstraint(Bo, [T \| R], S1, S2) :-<br>             S3 =.. [Bo, S1, T],<br>              aggrConstraint(Bo, R, S3, S2). |

Fig 4.22 *aggregate Prolog predicate*

Algorithm to convert ssRowsAggregate to Prolog equivalent

The algorithm in Fig 4.23 takes spreadsheet-specific *ssRowsAggregate* function and creates its SWI-Prolog equivalent. It is similar to the function in Fig 4.19, with the difference being the ROW-LIST-CREATION function being called during the processes, instead of the COLUMN-LIST-CREATION function.

---

SS-ROWS-AGGREGATE-TO-PROLOG
1. INPUT: *ssRowsAggregate ($B_0$,$C_{TL}$:$C_{BR}$,$R_0$,L)*
2. *L1* ← ROW-LIST-CREATION($C_{TL}$:$C_{BR}$)
3. *N* ← *length*(*L1*)
4. *L2* ← SETLEN(*L*, *N*)
5. OUTPUT: *subListAggregate($B_0$, L1, $R_0$, L2)*

---

Fig 4.23 ssRowsAggregate to Prolog Conversion

Algorithm to convert ssDiagonalAggregate to Prolog equivalent

The algorithm in Fig 4.24 takes spreadsheet-specific *ssDiagonalAggregate* function and creates its SWI-Prolog equivalent. It is similar to the function in Fig 4.19, with the difference being the DIAGONAL-LIST-CREATION function being called during the processes, instead of the COLUMN-LIST-CREATION function.

---

SS-DIAGONAL-AGGREGATE-TO-PROLOG
1. INPUT: *ssDiagonalAggregate ($B_0$,$C_{TL}$:$C_{BR}$,$R_0$,L)*
2. *L1* ← DIAGONAL-LIST-CREATION($C_{TL}$:$C_{BR}$)
3. *N* ← *length*(*L1*)
4. *L2* ← SETLEN(*L*, *N*)
5. OUTPUT: *subListAggregate($B_0$, L1, $R_0$, L2)*

---

Fig 4.24 ssDiagonalAggregate to Prolog Conversion

Algorithm to convert ssBackDiagonalAggregate to Prolog equivalent

The algorithm in Fig 4.25 takes spreadsheet-specific *ssBackDiagonalAggregate* function and creates its SWI-Prolog equivalent. It is similar to the function in Fig 4.19, with the difference being the BACK-DIAGONAL-LIST-CREATION function being called during the processes, instead of the COLUMN-LIST-CREATION function.

---

SS-BACK-DIAGONAL-AGGREGATE-TO-PROLOG
1. INPUT: $ssBackDiagonalAggregate\ (B_0, C_{TL}{:}C_{BR}, R_0, L)$
2. $L1 \leftarrow$ BACK-DIAGONAL-LIST-CREATION($C_{TL}{:}C_{BR}$)
3. $N \leftarrow length(L1)$
4. $L2 \leftarrow$ SETLEN($L, N$)
5. OUTPUT: $subListAggregate(B_0,\ L1,\ R_0,\ L2)$

---

Fig 4.25 ssBackDiagonalAggregate to Prolog Conversion

Algorithm to convert ssPairCellsAggregate to Prolog equivalent

The algorithm in Fig 4.26 takes spreadsheet spreadsheet-specific *ssPairCellsAggregate* function and creates its SWI-Prolog equivalent expression.

---

SS-PAIR-CELLS-AGGREGATE-TO-PROLOG
1. INPUT: $ssPairCellssAggregate\ (C_{TL1}{:}C_{BR1}\ ,\ B_0\ ,\ C_{TL2}{:}C_{BR2}\ ,\ R_0\ ,\ L)$
2. $L1 \leftarrow$ LIST-CREATION ($C_{TL1}{:}C_{BR1}$)
3. $L2 \leftarrow$ LIST-CREATION ($C_{TL2}{:}C_{BR2}$)
4. **if** $length(L) \neq length(L2)$ **then**
5. OUTPUT *Show-Error-Msg()*
6. **else**
7. $N \leftarrow length(L1)$
8. $L3 \leftarrow$ SETLEN($L, N$)
9. OUTPUT: $pairsAggregate(L1, B_0,\ L2,\ R_0,\ L3)$

---

Fig 4.26 ssPairCellsAggregate to Prolog Conversion

The function then proceeds as follows:

- It takes the *ssPairCellsAggregate* function as an input, creates two list *L1* and *L2*, using the LIST-CREATION function with the matrices $C_{TL1}{:}C_{BR1}$ and $C_{TL2}{:}C_{BR2}$ as arguments. It then checks to see if both the lengths of the lists are equal, and outputs the *pairsAggregate* prolog predicate defined in Fig 4.27. It lists lengths are not equal it outputs an error message.

| Prolog Predicate located in the Excel PL library |
|---|
| pairsAggregate([],_,[],_,_). |
| pairsAggregate([F1| R1], B0, [F2| R2], R0, [F3 | R3]) :- |
| T1 = .. [B0,F1,F2], |
| T2 = .. [R0,T1,F3], |
| T2, |
| pairsAggregate(R1,B0,R2, R0,R3). |

Fig 4.27 *pairsAggregate Prolog predicate*

The *pairsAggregate* prolog predicate is used to recursively format the ssPairCellsAggregate spreadsheet-specific function into the appropriate prolog syntax.

Algorithm to convert ssMin to Prolog equivalent

The algorithm in Fig 4.28 takes spreadsheet spreadsheet-specific *ssMin* function and creates its SWI-Prolog equivalent expression.

---

SS-MIN -TO-PROLOG
1. INPUT: *ssMin (C)*
2. **if** *IsSingleValue*(C) **then**
3.     OUTPUT *min(C)*
4. **else**
5.     OUTPUT *Show-Error-Msg()*

---

Fig 4.28 ssMin to Prolog Conversion

The function then proceeds as follows:

- This function takes the spreadsheet-specific function *ssMin(C)*. It then checks to

  see if *C* is a single value, if it is then it outputs min(C), otherwise it will show an

  error message as shown on **Line 5**.

Algorithm to convert ssMax to Prolog equivalent

The algorithm in Fig 4.29 takes spreadsheet spreadsheet-specific *ssMax* function and creates its SWI-Prolog equivalent expression.

---

SS-MAX -TO-PROLOG
1. INPUT: *ssMax (C)*
2. **if** *IsSingleValue*(C) **then**
3.     OUTPUT *max(C)*
4. **else**
5.     OUTPUT *Show-Error-Msg()*

---

Fig 4.29 ssMax to Prolog Conversion

The function then proceeds as follows:

- This function is similar to the *ssMin* function with the difference being that it
  outputs the *max(C)* on **Line 3**, instead of *min(C)*.

Prolog Predicate nthElement

The *nthElement(N,VList,V)* prolog predicate states that the constraints *V* needs to be equal to the $N^{th}$ variable (starting from 1) of the list *VList. VList* can contain integers.

| Prolog Predicate located in the Excel PL library |
|---|
| nthElement(N, VList, V) :-<br>    length(VList, L),length(VB, L),<br>    VB in 0..1, sumExp(VB, S),<br>    S #= 1, length(Index, L),<br>    setIndex(Index, 1, L),<br>    multiSum(Index, VB, MS),<br>    MS #= N,<br>    multiSum(VList, VB, MS1),<br>    MS1 #= V.<br><br>multiSum([], _, 0) :- !. |

```
multiSum(_, [], 0) :- !.
multiSum([A|L1], [B|L2], A * B + MS) :-
                     multiSum(L1, L2, MS).

setIndex([], L, U) :-  L #> U, !.
setIndex([L], L, L) :- !.
setIndex([L | Index], L, U) :-  L1 #= L + 1,
                       setIndex(Index, L1, U).

sumExp([], 0).
sumExp([V | VL], V + S) :-  sumExp(VL, S).
```

## 4.3.2.4 Framework required Spreadsheet-Specific Transformations

This section describes the algorithms that will be used to convert framework required the spreadsheet-specific constraint functions into their SWI-Prolog equivalent. The two framework required spreadsheet-specific functions discussed in section 4.2.2.3 are *ssVarRanges*() which is used to specify variable ranges, and *ssConstraintRanges*() used to specify the constraint ranges.

Algorithm to compile CLP program using ssVarRanges and ssConstraintRanges

The algorithm in Fig 4.30 takes *ssVarRanges(C1)* and *ssConstraintRanges(C2)* as input and build a CLP program.

---

SS-COMPILE-TO-PROLOG
1. INPUT: *ssVarRanges(C1)*
2. INPUT: *ssConstraintRanges(C2)*
3. *S ← { }   ; a sequence of Prolog constraints*
4.
5. *% Variable Range section*
6. *L1 ←* LIST-CREATION(*C1*)
7.     **foreach** cell variable *V* **in** *L1*
8.         **if** *V.value ≠ EmptyString* **then**
9.             *cs ←* DOMAIN-TRANSFORMATION(*V.value*)

10.       $S \leftarrow$ append($cs$, $S$)

11.

12.  *% Constraint Range section*

13.  *L2* $\leftarrow$ LIST-CREATION(*C2*)

14.    **foreach** cell variable *V* **in** *L2*

15.      **switch** *V.value*

16.        **case** *ssDomain(C, Min, Max)***:**

17.            cs $\leftarrow$ SS-DOMAIN-TO-PROLOG(*V.value*)

18.      **case** *ssAllDifferent(C)***:**

19.            cs $\leftarrow$ SS-ALL-DIFFERENT-TO-PROLOG (*V.value*)

20.      **case** *ssColsAllDifferent(C)***:**

21.            cs $\leftarrow$ SS-COLS-ALL-DIFFERENT-TO-PROLOG (*V.value*)

22.      **case** *ssRowsAllDifferent(C)***:**

23.            cs $\leftarrow$ SS-ROWS-ALL-DIFFERENT-TO-PROLOG (*V.value*)

24.      **case** *ssColsAggregate ($B_0$,$C_{TL}$:$C_{BR}$,$R_0$,L)***:**

25.            cs $\leftarrow$ SS-COLS-AGGREGATE -TO-PROLOG (*V.value*)

26.      **case** *ssRowsAggregate ($B_0$,$C_{TL}$:$C_{BR}$,$R_0$,L)***:**

27.            cs $\leftarrow$ SS-ROWS-AGGREGATE -TO-PROLOG (*V.value*)

28.      **case** *ssDiagonalAggregate ($B_0$,$C_{TL}$:$C_{BR}$,$R_0$,L)***:**

29.            cs $\leftarrow$ SS-DIAGONAL-AGGREGATE -TO-PROLOG (*V.value*)

30.      **case** *ssBackDiagonalAggregate ($B_0$,$C_{TL}$:$C_{BR}$,$R_0$,L)***:**

31.            cs $\leftarrow$ SS-BACK-DIAGONAL-AGGREGATE -TO-PROLOG (*V.value*)

32.      **case** *ssPairCellsAggregate($C_{TL1}$:$C_{BR1}$, $B_0$,$C_{TL2}$:$C_{BR2}$,$R_0$,L)***:**

33.            cs $\leftarrow$ SS-PAIR-CELLS-AGGREGATE -TO-PROLOG (*V.value*)

34.      **case** *nthElement(N,Vlist,V)***:**

35.            *L3* $\leftarrow$ LIST-CREATION(*Vlist*)

36.            cs $\leftarrow$ nthElement(N,*L3*,V)

37.      **case** *ssMIN(C)***:**

38.            *S1* $\leftarrow$ SS-MIN -TO-PROLOG (*V.value*)

39.      **case** *ssMAX(C)***:**

40.            *S1* $\leftarrow$ SS-MAX-TO-PROLOG (*V.value*)

41.    $S \leftarrow$ append(*cs*, *S*)

42.

43.  *% Construct the CLP program in the form of:*

44.  *Program* $\leftarrow$ {*:- use_module(library(bounds)).*}

45.  *Program* $\leftarrow$ append(*:- use_module(library(excel)). , Program*)

46.  *Program* $\leftarrow$ append(*mainQuery(L1) :-, Program*)

47.  *Program* $\leftarrow$ append(*<S>, Program*)

48.  *Program* $\leftarrow$ append( *labeling([<S1>], L1)., Program*)

49.  OUTPUT: *Program*

Fig 4.30 Compile CLP program using ssVarRanges and ssConstraintRanges as input

The function then proceeds as follows:

- The function takes *ssVarRanges(C1)* and *ssConstraintRanges(C2)* as input, then instantiates a variable *S*, which will be used to store a sequence of prolog constraints.

- **Line 6** sends the contents *C1* of the *ssVarRange(C1)* function to the LIST-CREATION FUNCTION, and returns a list *L1*. We then iterate through the list L1 starting in **Line 7**, and if the cell variable *V* in *L1* has a value, we send it's value to the DOMAIN-TRANSFORMATION function, and append the result *cs* to the back of *S* to create a partial list of constraints separated by commas.

- We then proceed to build a complete list of constraint, by passing the *C2* input variable to *ssConstraintRanges(C2)* to the LIST-CREATION(C2) function, which returns a list of cell values *L2*. Starting in **Line 14**, we proceed to check the cell variable *V* in *L2* for matching constraint value, by passing *V.Value* through a case statement.

- If a matching spreadsheet-specific function is found, the cell variable *V*'s value is passed to it, with the result cs being appended to the sequence of variable *S*. The exception to the rule occurs in **Lines 38** and **40** where the result of either the SS-MIN-TO-PROLOG or SS-MAX-TO-PROLOG are assigned to the variable *S1*, because that will be used as the parameter for the labeling function on **Line 48**.

- Another exception to the rule occurs on **Line 35**, where the 2nd parameter of the *nthElement(N,Vlist,V) Vlist* is sent to the LIST-CREATION function. Its result *L3* is appended back to the *nthElement* as *nthElement(N,L3,V)*, which is then assigned to the *cs* variable. The *cs* variable is then used to append to the end of variable *S*'s sequence of Prolog constraints .

- The final step involves constructing the constraint logic program beginning by appending the 2 headers needed by SWI-Prolog, one of which is used to process CLP(FD) programs as shown in **Line 44** and the custom excel library module shown in **Line 45**.

- The last part of the program involves the following sequence of events starting on **Line 46**:

  1. Taking the variable range list L1 and combining it with mainQuery(L1) and as part of the labeling([<S1>,L1).

  2. Taking the list of constraints S and using it as part of the body of the program (*mainQuery(L1) :- <S>, labeling([<S1>], L1).*) .

- The resulting output will have the format shown in the table below:

```
:- use_module(library(bounds)).
:- use_module(library(excel)).

mainQuery(L1):-
                    <S>,
                    labeling([<S1>],L1).
```

### 4.3.3 CHR/Excel Pl Library User Interface

The CSeE system allows advanced users (e.g. users with prolog/CLP programming experience) to add to the Excel prolog library or build constraint handling rules through a Windows Form activated through an Add-In menu at the Spreadsheet Layer. The new library addition is then compiled and run on the SWI-Prolog constraint solver, through the .NET Wrapper. Once the addition is complete, the users can simply reference that prolog predicate in the Spreadsheet and the Parser/Analyzer will add it to the new CLP program.

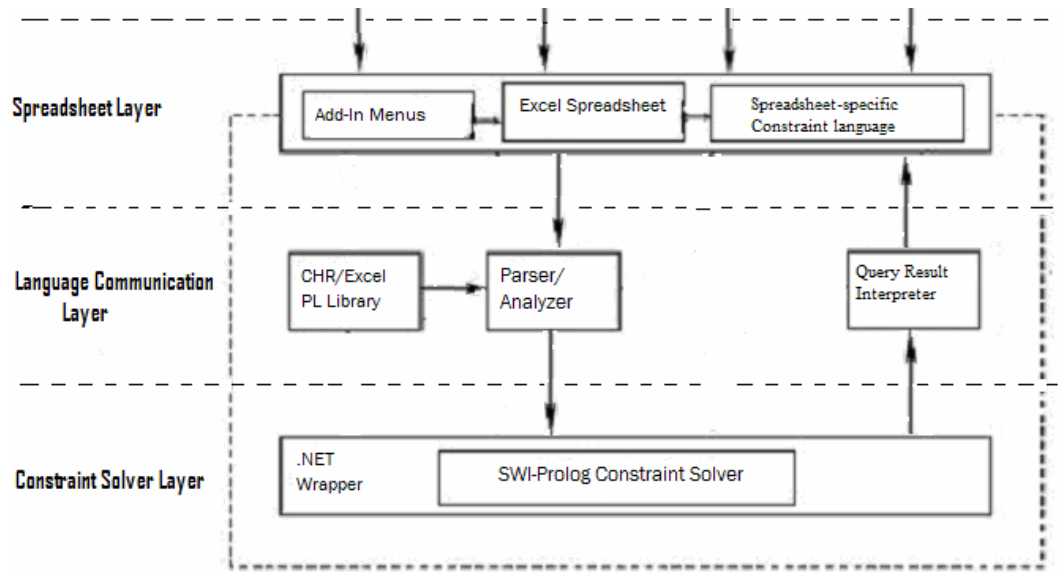

Fig 4.31 CHR/Excel PL library addition

## 4.4 Constraint Solver Layer

This layer is comprised of the .NET Wrapper and a 3[rd] party constraint solver, which in this case is the SWI-Prolog.

### 4.4.1 .NET Wrapper

Rich class libraries, templates, languages and technologies make the .NET framework an ideal platform for developing both the communication interface and Office applications.  In addition to reusability, the framework allows developers to use a multitude of languages such as C# or VB.NET to manipulate the Office application's object model, and interface with third party software in a seamless managed environment [9].

One such library provided by the .NET Framework's is the System Diagnostics library which is used to wrap SWI-Prolog's command window. The library provides functionality to send data (e.g. programs generated from the parser, queries) and receive data (e.g. results from the parser) from SWI-Prolog's command window, in a synchronous mode. It also allows the SWI-Prolog functionality to be ported to other .NET applications.

## 4.4.2 SWI –Prolog

The SWI-Prolog was developed in 1986 and is based on a subset of the Warren Abstract Machine (WAM). It provides a simple constraint solver which is a subset of the SICStus CLP (FD) syntax which requires that it is explicitly loaded using the following library call (e.g.,*:- use_module(library(bounds))*.) [19].

In addition, it provides a way to write constraint solvers through the use of CLP (Constraint Logic Programs), particularly application-specific constraints such as scheduling [19].

# CHAPTER 5

# CSeE System Implementation

The CSeE system is a COM Add-In for Excel 2003 developed using the C# language and Visual Studio Tools for Office 2005. In this chapter I will go through showing CSeE system's implementation details ranging from creating a COM Add-in to code snippets of the overall architecture.

## 5.1 Add-In Enhancements

Microsoft Office's application has a COM object model and the generic interface like IDTExensibility2 allows a "Shared Add-in" to be used from any selected Office application. In the past creating a COM add-in had proved to be very tricky, but Visual Studio has provided a wizard named "Shared Add-in" to assist in the creation process. The steps in creating a "Shared Add-in" are described below:

1. The Visual Studio provided "Shared Add-in" can be found under Other Project Types > Extensibility > Shared Add-in as shown in the figure below.

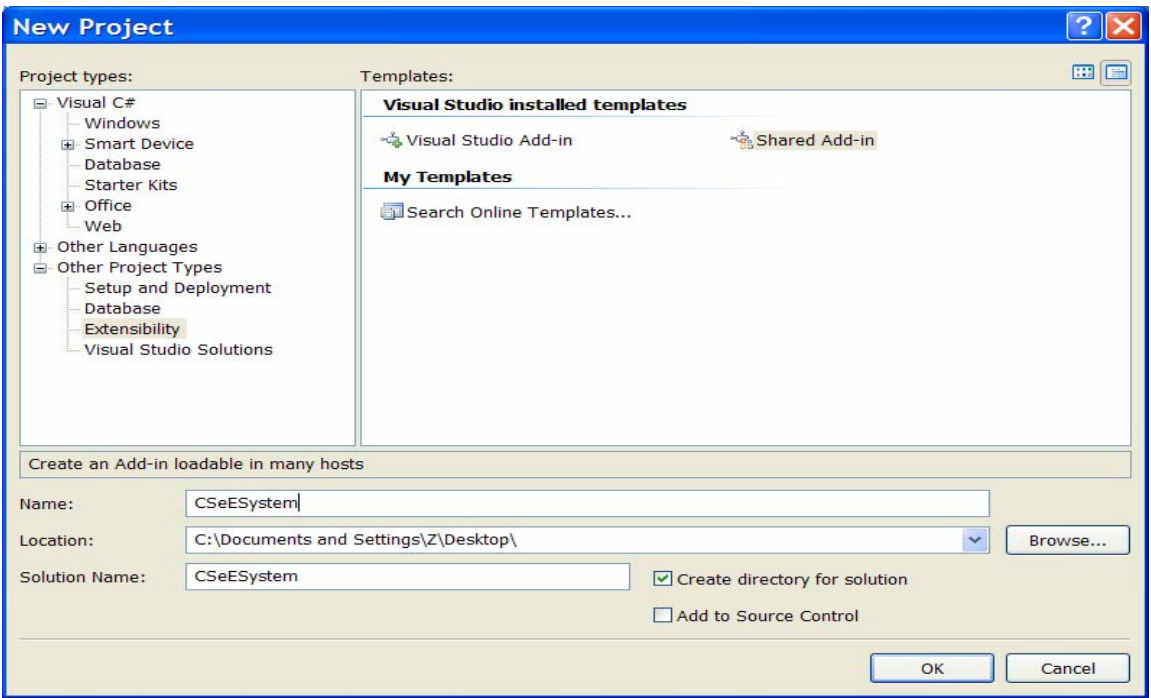

Fig 5.1: Creating the CSeE COM Add-in

2.  Once you have selected the "Shared Add-in" template and entered a name, the next step will be to select the programming language which will be C# as shown in the figure below.

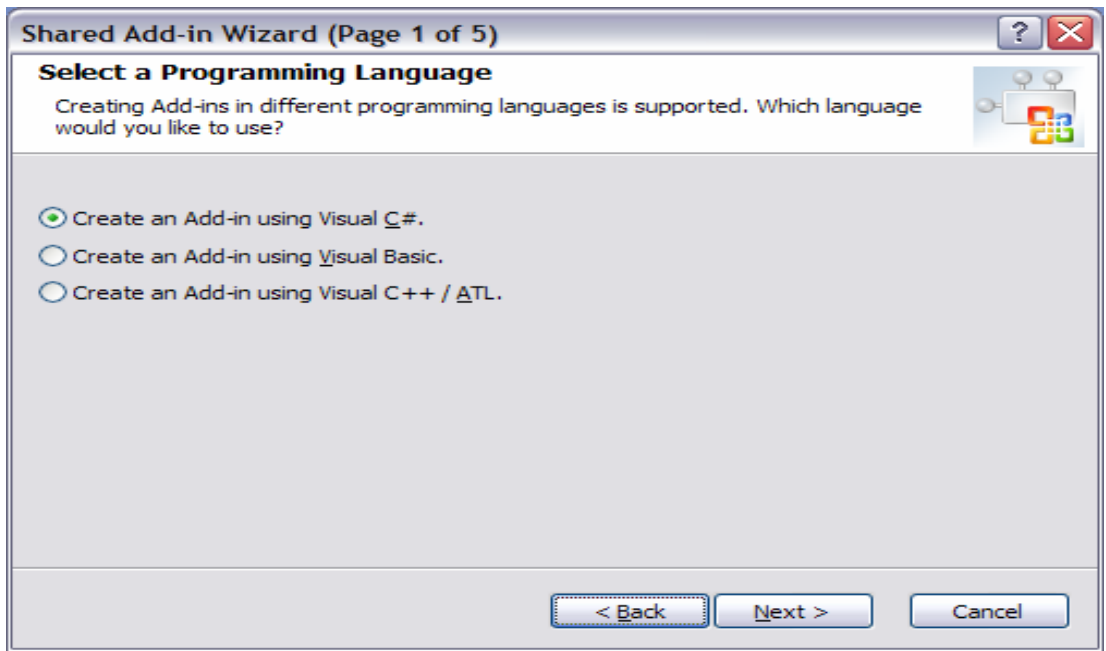

Fig 5.2: Selecting the Add-in Programming Language

3. The next step will be to select the application host which will be Microsoft Excel, shown in the figure below.

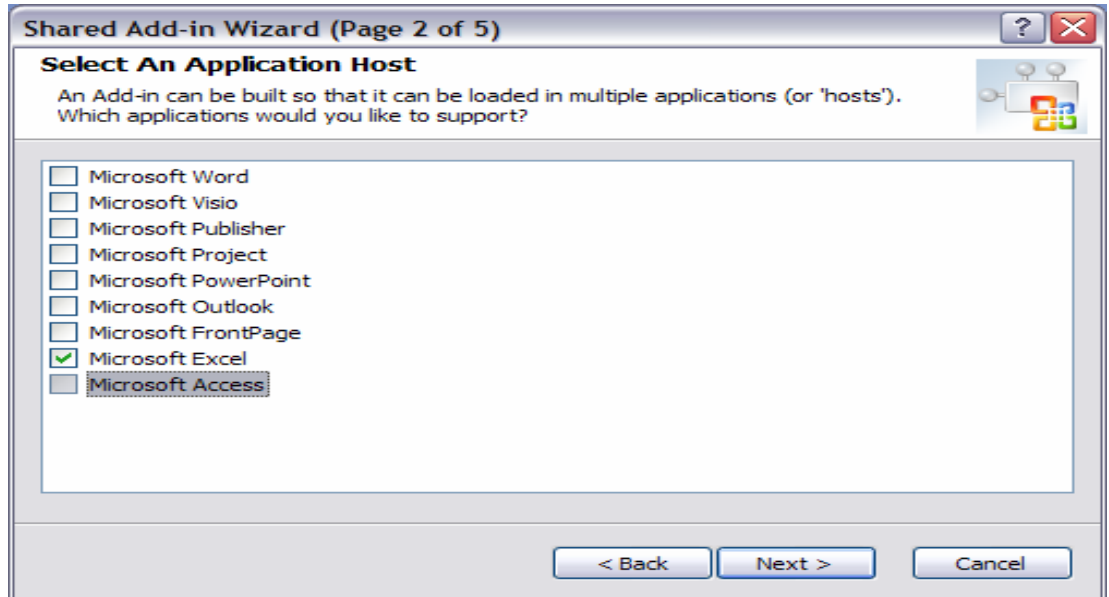

Fig 5.3: Selecting the application host

4. The next step requires that the Add-in be given a name and a brief description for registry setup; this can be seen in the figure 5.4 below.

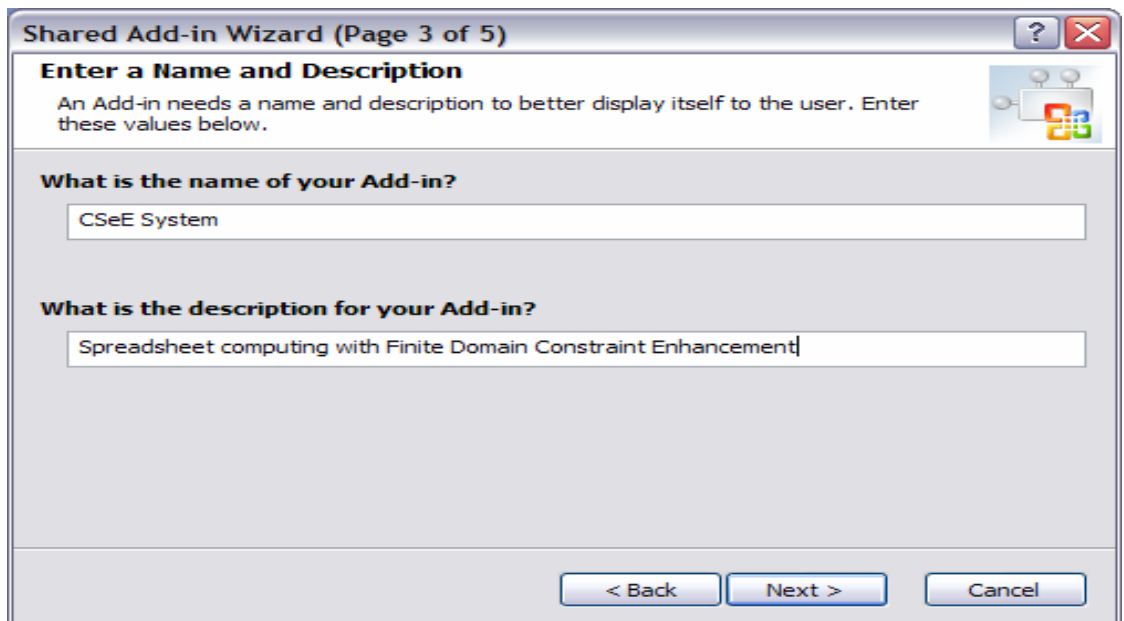

Fig 5.4: Giving the name and description required for the registry

5. The last setup as shown in the figure below involves determining how the COM
   add-in will be loaded in the registry under the current user key
   HKEY_CURRENT_USER or under the local machine key
   HKEY_CURRENT_MACHINE which will make the COM Add-in available to
   all the users on the machine.

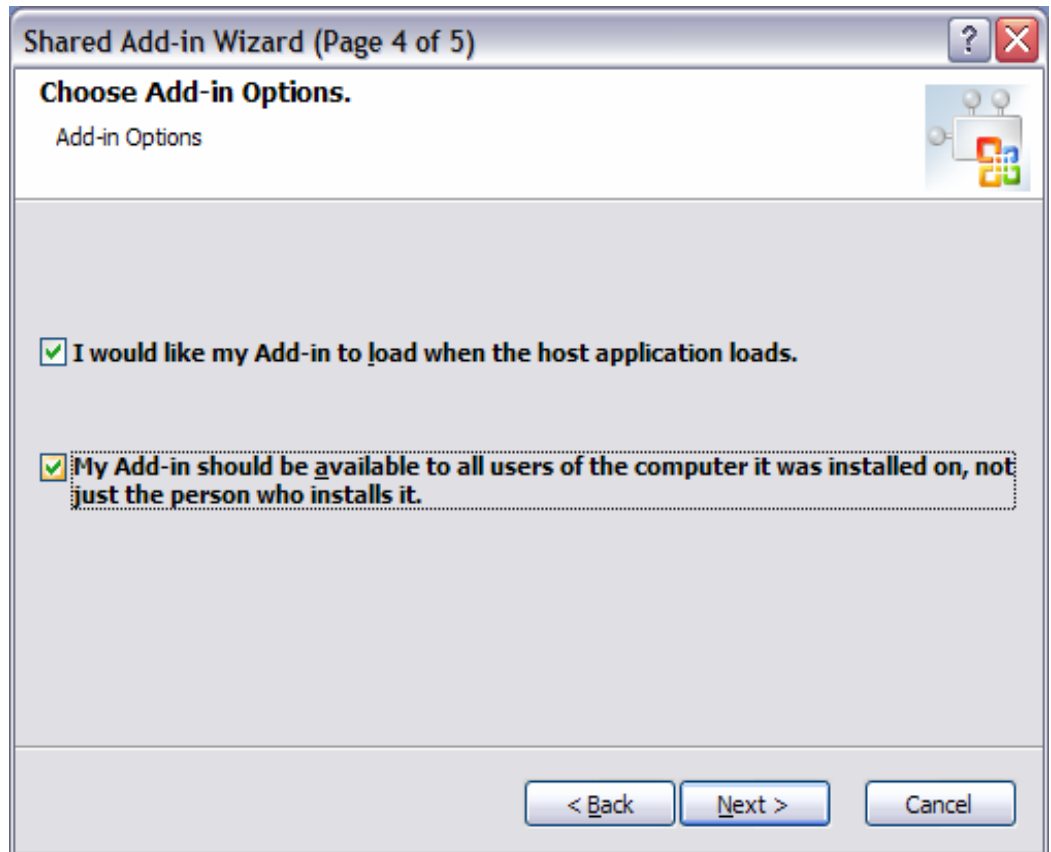

Fig 5.5: Choosing how the COM Add-in will be loaded

6. The next step will be setting the external start programs. You can setup the
   external start program by right clicking on the project, then going to the debug tab
   and in the start action section enter the following *"C:\Program Files\Microsoft
   Office\OFFICE11\EXCEL.EXE"*.  This would allow the code to be stepped
   through.

7. The last step will be setting up references for the desired libraries as shown in the figure below.

```
using System;
using System.Data;
using System.Drawing;
using Extensibility;
using System.Runtime.InteropServices;
using Office = Microsoft.Office.Core;
using Excel = Microsoft.Office.Interop.Excel;
using Microsoft.VisualStudio.Tools.Applications.Runtime;
using System.Windows.Forms;
using System.Collections.Generic;
using System.IO;
```

Fig 5.6: Namespaces used in the solution

## 5.1.1 OnConnection Method

The OnConnection gets notified when the Add-in is being loaded. So we will instantiate the application object and the COM Add-in instance shown in the figure below.

```
private Microsoft.Office.Interop.Excel._Application appObj;
private Microsoft.Office.Core.COMAddIn addInInstance;

public void OnConnection(object application,
                         Extensibility.ext_ConnectMode connectMode,
                         object addInInst, ref System.Array custom)
{
    appObj = (Microsoft.Office.Interop.Excel.Application)application;

    addInInstance = (Microsoft.Office.Core.COMAddIn)addInInst;

}
```

Fig 5.7: the OnConnection Method

## 5.1.2 The On Startup Complete Method

This is a function which is called when the host application has completed loading. As shown in the figure below.

```
private Office.CommandBar menuBar;
private Office.CommandBarButton menuParse;
private Office.CommandBarPopup solutionSubMenuGroupButton;
private Office.CommandBarButton submenuNextSolution;
private Office.CommandBarButton submenuPreviousSolution;
private Office.CommandBarButton menuOriginalState;
private Office.CommandBarButton menuExcelLib;
private Office.CommandBarButton menuHelp;
Office.CommandBarPopup cmdBarControl = null;

public void OnStartupComplete(ref System.Array custom)
{
   //Check directory existens and create sub-directories if they doe not exist
    this.GetDirectory();

    //Setup the Menu Bar
    this.InitiateTheMenuBar("C&SeE System");

    //Then setup the menus,sub-menus,etc..
    if (cmdBarControl != null)
    {
       this.CreateMenusAndSubMenus();

       //Set the SheetDirty indexes to false
       for (int i = 0; i < 20; i++)
        sheetDirty[i] = false;
     }

}
```

Fig 5.8: The OnStartUpComplete Method

The method proceeds as follows:

1. It calls the `this.GetDirectory()` function, whose responsibility is to check if the directory *MyCLPAddIN* and *MyCLPAddIN/Help* exist under the *Program Files\pl* folder, and if they do not then create them.

2. It then proceeds to create the main menu called "CSeE System". The menu is created by adding an Office.CommandBarPopup to the menu bar, and then you proceed to add the desired caption [10, 11]. The code within the function is shown below.

```csharp
private void InitiateTheMenuBar(string menuCaption)

{
  try
  {
    //Add CommandBarPopup to the ActiveMenuBar
    menuBar = (Office.CommandBar)appObj.CommandBars.ActiveMenuBar;
    cmdBarControl = (Office.CommandBarPopup)menuBar.Controls.Add(
                        Office.MsoControlType.msoControlPopup,
                    missing,missing,        menuBar.Controls.Count, true);
    //displays the name of the menu item
    cmdBarControl.Caption = menuCaption;
  }
  catch (Exception ex)
  {
    MessageBox.Show(ex.Message, ex.Source, MessageBoxButtons.OK,
    MessageBoxIcon.Error);
  }
}
```

Fig 5.9: Code to add a menu named CSeE System AddIn

3. Once the menu has been created without any errors, then the next setup involves creating the submenus. This requires creating sub-menu groups and sub-menu buttons in addition to adding click event handlers as well as icons [10, 11]. Figure 5.10 shows the partial code necessary to implement the sub-menus and figure 5.11 shows the end result.

```csharp
private void CreateMenusAndSubMenus()
{
  menuParse = CreateMenuButton("&ParseBuild", 65);
  menuParse.Click += new

                    Microsoft.Office.Core._CommandBarButtonEvents_ClickEventHan
                    dler(MenuParse_Click);

  //Solution Option sub-menu group and event handler
  solutionSubMenuGroupButton = CreateSubMenuGroupButton("&Solution Options",
true);
  solutionSubMenuGroupButton.Enabled = false;
        submenuNextSolution =
      CreateSubMenuButtons(solutionSubMenuGroupButton, "&Next  Solution",
      3006);

submenuNextSolution.Click +=new
      Microsoft.Office.Core._CommandBarButtonEvents_ClickEventHandler(submenuNe
xtSolution_Click);
  submenuNextSolution.Enabled = false;

  submenuPreviousSolution = CreateSubMenuButtons(solutionSubMenuGroupButton,
"&Previous Solution", 3005);
  submenuPreviousSolution.Click +=
              new
Microsoft.Office.Core._CommandBarButtonEvents_ClickEventHandler(submenuPrevious
Solution_Click);
```

```
   submenuPreviousSolution.Enabled = false;

   menuOriginalState = CreateMenuButton("&Original State", 3007);
   menuOriginalState.Click +=
                           new
Microsoft.Office.Core._CommandBarButtonEvents_ClickEventHandler(menuOriginalSta
te_Click);
   menuOriginalState.Enabled = false;

   menuExcelLib = CreateMenuButton(@"Add to CHR\Excel Lib",0);
   menuExcelLib.Click +=
                           new
       Microsoft.Office.Core._CommandBarButtonEvents_ClickEventHandler(menuExcelLib_Click
           );
   menuExcelLib.Enabled = true;

   menuHelp = CreateMenuButton(@"CSeE Help", 0);
   menuHelp.Click +=
               new
       Microsoft.Office.Core._CommandBarButtonEvents_ClickEventHandler(menuHelp_Click);
   menuHelp.Enabled = true;
}
```

Fig 5.10: Partial Sub Menu and Sub Menu group creation code

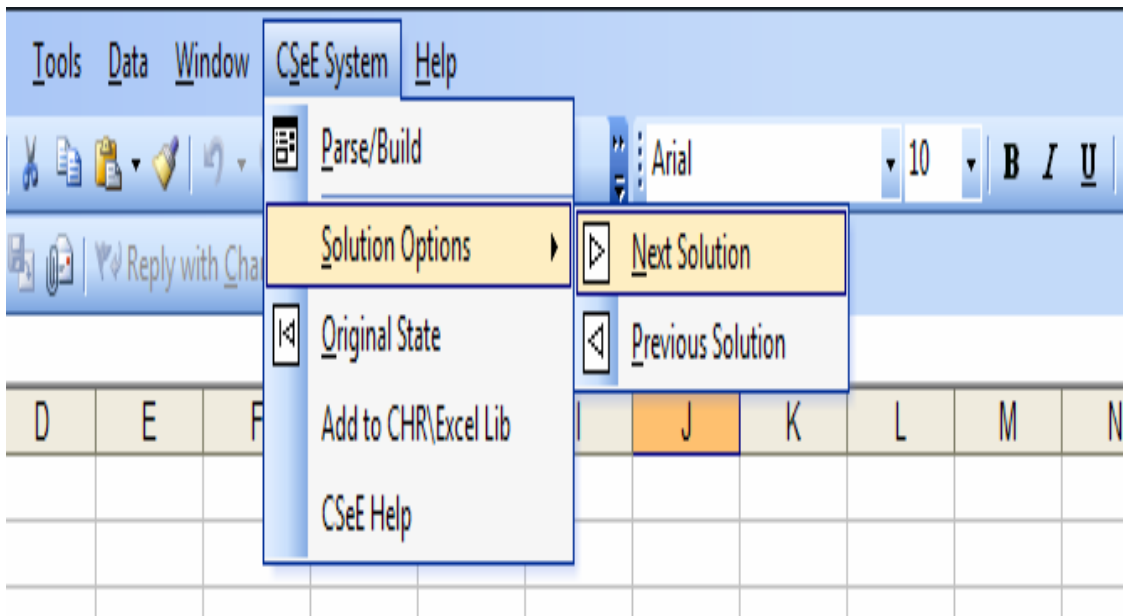

Fig 5.11: Screen shot of Menu, sub menu groups and sub menus

## 5.2 Spreadsheet-Specific Functions Enhancement

When adding the customized functions such as spreadsheet-specific functions, the use of the Excel function formula is useful since it allows users to drag the function arguments (e.g. variable range information arguments located in the A1:I9 cell matrix shown in the figure below.

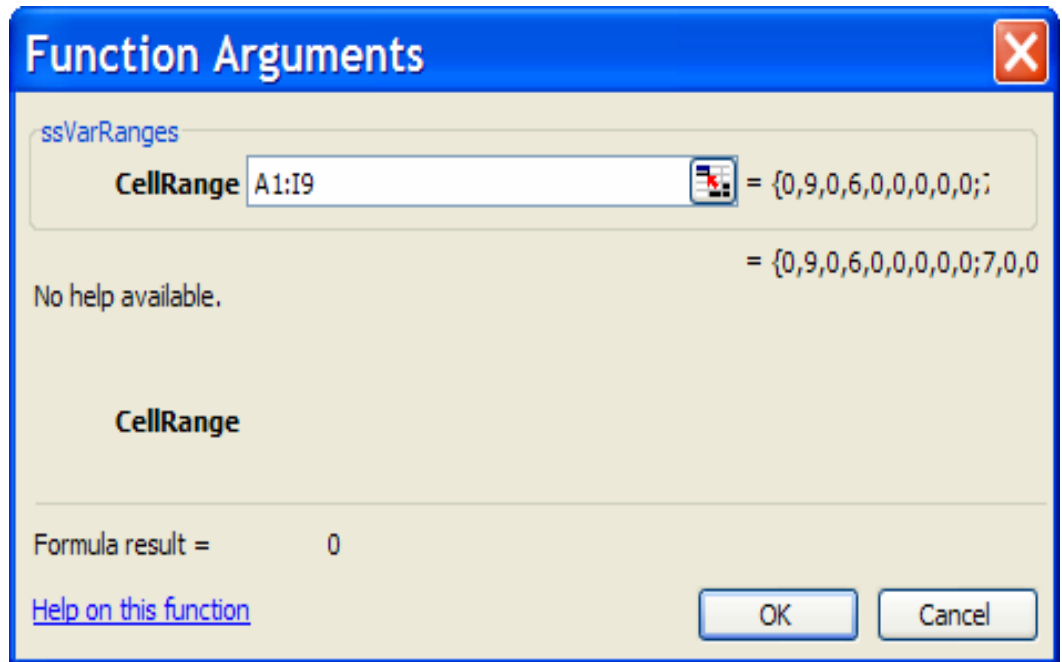

Fig 5.12 setting ssVarRanges arguments using the customized functions

## 5.2.1 Automation Add-in Function Class

Excel 2003 allows one to write customized functions to add Excel Formulas by using a customization technology known as automation add-in [9]. You can create a C# class containing customized functions and you need to implement *RegisterFunction* and *UnregisterFunction*, functions. These functions allow the Excel application to treat them as an automation add-in [9]. The figure below shows the CSeE system's implemented *Function* class with a partial listing of the functions.

```csharp
[ClassInterface(ClassInterfaceType.AutoDual), ComVisible(true)]
public class Functions
{
  public Functions(){}

  public object ssVarRanges(object cellRange)
```

```
{
    return cellRange;
}

public object ssConstraintRanges(object cellRange)
{
    return cellRange;
}

[ComRegisterFunctionAttribute]
public static void RegisterFunction(Type type)
{
  Registry.ClassesRoot.CreateSubKey(GetSubKeyName(type));
}

[ComUnregisterFunctionAttribute]
public static void UnregisterFunction(Type type)
{
  Registry.ClassesRoot.DeleteSubKey(GetSubKeyName(type), false);
}

private static string GetSubKeyName(Type type)
{
        string s = @"CLSID\{" + type.GUID.ToString().ToUpper() +
    @"}\Programmable";
    return s;
}
}
```

Fig 5.13 automation Add-in function class

## 5.2.2 Required function search

The figure below shows the necessary code to search for a spreadsheet-specific required function, which in this case it the search for the *ssConstraintRanges* function. There is also one that performs the search for the other spreadsheet-specific required function *ssVarRanges*.

```
private string FindConstraints()
{
    //Constraint get range from 1st cell
    Excel.Range range1 =
    ((Excel.Worksheet)appObj.ActiveSheet).get_Range("$A1", missing);
```

```
//Search for that ssConstraintRanges key word
Excel.Range foundRange =
                    range1.Find(
                    "ssConstraintRanges",
                    range1.get_Item(1, 1),
                    missing,
                    Excel.XlLookAt.xlPart,
                    missing,
                    Excel.XlSearchDirection.xlNext,
                    missing,
                    missing,
                    missing);

//If the function is found return a the value, else the error
 if (foundRange != null)
 {
    return foundRange.Value2.ToString();
 }
 else
 {
    return "Error";
 }
}
```

Fig 5.14 Find Constraints function search

# 5.3 Menu, Sub-menu event handling functions

The section covers some aspects of the menu, sub-menu event handler functions and also shows the code that is used to set the worksheet index and ranges list.

## 5.3.1 Parse Menu event handling function

The figure below shows the code snippet used to build the constraint logic program described in chapter 4's figure 4.30. In addition it calls the *GenerateSolutions()* function, which finds all the possible solutions and displays the 1[st] solution in the appropriate cells.

```csharp
private void MenuParse_Click(Office.CommandBarButton ctrl, ref bool CancelDefault)
{
    String[] rangeLists;

    foreach (Excel.Worksheet sheet in appObj.Worksheets)
    {
        //Set the worksheet name and index
        sheetNames.Add(new WorkSheetNameIndex(sheet.Index, sheet.Name));
    }

    //Find and get the ssVarRanges() function and return its argument
    string domainVarRanges = this.FindDomainVariables();

    if (domainVarRanges.ToUpper() != "ERROR")
    {
        //Find and get the ssConstraintRanges() function and return its argument
        string constraintVarRanges = this.FindConstraints();

        if (constraintVarRanges.ToUpper() != "ERROR")
        {
            //Sends A1:B2 for sheet 1 and returns 1-A1:B2;
            //For multiple sheets eg. sheet 1 A1:B2 and sheet 2 C1:F3 it retuns 1-A1:B2;2-C1:F3
            rangeLists = this.SetWorksheetIndexAndRangeList(domainVarRanges).Split(';');

            Variable....

            Contraints Section ................................................................

            File Write section .......................................................

            #region Generate and write 1st solution to the cell

              this.GenerateSolutions();

            #endregion

        }
        else
            this.ConstraintErrorMessage();
    }
    else
        this.DomainErrorMessage();

}
```

Fig 5.15 Parse Menu event handler function

## 5.3.2 The Set Worksheet Index and Range List helper function

The figure below shows the code snippet used to append the sheet indices to the range lists and is used in the parse menu event handler function above. Its purpose is to keep track of which sheet has the appropriate variable information.

So when a large problem is encountered, it can be broken down into multiple sheets, to aid in handling its complexity (e.g. for sheet 1 with a range of A1:B2, it returns 1-A1:B2, for multiple sheets it would return *Sheetindex1*- CellMatrix**;** *SheetIndex2*- CellMatrix).

```csharp
private string SetWorksheetIndexAndRangeList(string rangeAndSheetNames)
{
    string[] rangeList;
    //=ssVarRanges(F31:G31,Sheet2!D24,Sheet2!E29:F29)
    if(rangeAndSheetNames.StartsWith("="))
        rangeAndSheetNames = rangeAndSheetNames.Remove(0,1);

    rangeAndSheetNames = rangeAndSheetNames.Replace(@"ssVarRanges(", "").Replace(")", "");
    rangeAndSheetNames = rangeAndSheetNames.Replace(@"ssConstraintRanges(", "").Replace(")", "");

    if(rangeAndSheetNames.Contains(":"))
        rangeList = rangeAndSheetNames.Split(',');
    else
        rangeList = rangeAndSheetNames.Split(';');
    string returningList = "";

    for(int i = 0; i < rangeList.Length; i++)
    {
        int index = this.SeeIfMultipleSheetRange(rangeList[i]);
        if(index != -1)
            returningList += rangeList[i].ToString().Replace(sheetNames[index -1].SheetName + "!",
                                      index.ToString() + "-") + ";";
        else
        {
            if(rangeList[i].Substring(0) =="=",") //,A1:B2 makes it 1-A1:B2
                returningList += rangeList[i].Replace(",", "," + ReturnActiveSheetIndex() + "-") + ";";
            else //A1:B2 makes it 1-A1:B2
                returningList += ReturnActiveSheetIndex() + "-" + rangeList[i] + ";";
        }
    }

    return returningList;
}

private int SeeIfMultipleSheetRange(string rangeList)...

private string ReturnActiveSheetIndex()...
```

Fig 5.16 The Set Worksheet index and range list helper function

### 5.3.3 Remaining Menus event handling functions

The figure below displays the code snippets of the event handlers for the remaining menus with comments appearing before the function definition.

```csharp
//this function displays the next function starting from 2, if there is
//more than one solution, or just the only derived solution if there
// there is only 1 solution
private void submenuNextSolution_Click(
        Microsoft.Office.Core.CommandBarButton Ctrl, ref bool CancelDefault)
{
        this.GetNextSolution();
}

//Call the funtion to display the previous solution
private void submenuPreviousSolution_Click(
        Microsoft.Office.Core.CommandBarButton Ctrl, ref bool CancelDefault)
{
    this.GetPreviousSolution();
}

//On Click call the write previous solution function and enable the Next
// Solutions sub-menu
private void menuOriginalState_Click(
                        Office.CommandBarButton ctrl, ref bool CancelDefault)
{
        this.writepRESolutionsToCells();
        submenuNextSolution.Enabled = true;
}

//On Click display Add to Excel Library Interface
private void menuExcelLib_Click(
                        Office.CommandBarButton ctrl, ref bool CancelDefault)
{
    AddToExcelLibForm frm = new AddToExcelLibForm();
    frm.ShowDialog();
}

//On Click display Help Interface
private void menuHelp_Click(
                    Office.CommandBarButton ctrl, ref bool CancelDefault)
{
    frmHelp help = new frmHelp();
    help.ShowDialog();
}
```

Fig 5.17 Remaining Menu event handler functions

# 5.4 .CSeE System .NET Wrapper

The *CSeESysWrapper* is a class that wraps the SWI Prolog's *plcon.exe* window using the .NET 2.0's System.Diagnostics.Process namespace, thereby enabling the CSeE System to interface with SWI-Prolog. You begin interfacing with SWI-Prolog's in the following manner:

1. You begin by instantiating the class, and passing the location of the directory where the newly created CLP program is located, which in this case is the *MyCLPAddIN* located under the "Program Files/pl" directory.

2. This will instantiate a new process, assign the file name which is *plcon.exe*, redirect to the standard input, make sure *plcon.exe*'s window is hidden during execution, set the directory location, and start the process.

3. You would then compile the newly created CLP program by passing it to the *WriteToStandardinput()* function.

4. You would then send the query string "solve(L)." to the SendQueryGetResults() function, which will parse and write the query.

5. It would then send the directory location and GuiID (temp file) to the *OutputHandler()* function, which will sleep and wake up and check if the file is create.

6. Once the file is created it will sleep and wake up, until a response it received and the contents have been read.

7. It will delete the temp file, and return the solution(s) to the Spreadsheet Layer, where it will be reformatted to display in the appropriate spreadsheet cells.

The figure below shows the code of some of the steps described above, with the class instantiation and function calls shown above the class snippet.

```
//Class instantiation and function calls
private string SendToCSeESysWrapper()
{
    wrapper = new CSeESysWrapper(TempDirectory);
    wrapper.WriteToStandardInput(IniFile);
    string x = @"solve(L).";
    return wrapper.SendQueryGetResults(x, false);
}

//Class snippet
```

```csharp
public class CSeESysWrapper
{
    public Process process ;
    private string _dirLocation;
    private string _plcon = @"C:\Program Files\pl\bin\plcon.exe";

    public CSeESysWrapper(string dirLocation)...
    public void WriteToStandardInput(string file)...
    public string SendQueryGetResults(string query)...

    private string OutputHandler(string fileName)
    {
        StreamReader strreader = null;
        string output = string.Empty;
        bool outputDone = false;

        while (!outputDone)
        {

            while (!File.Exists(fileName))
            {
                System.Threading.Thread.Sleep(10);
            }
            System.Threading.Thread.Sleep(10);

            try
            {
                strreader = new StreamReader(fileName);
                output = strreader.ReadToEnd();
                strreader.Close();
                outputDone = true;
            }
            catch
            {
                continue;
            }
            try
            {
                //remove temp file
                File.Delete(fileName);
            }
            catch { }
        }
        return output;//returns to the spreadsheet Layer [[1,2,3,4]]
    }
}
```

Figure 5.18: The CSeESysWrapper function call and class code snippet

# CHAPTER 6

## Examples and Results

This chapter demonstrates how the CSeE system allows the general user to extend the spreadsheet's ability to solve constraint satisfaction problems (e.g. puzzle solving, resource allocation problems, and scheduling problems). These constraint satisfaction problems were previously time consuming, arduous or even inaccessible due to semantic and syntactic complexities.

The examples include popular recreational puzzles namely Sudoku and N-Queens, and a resource allocation problem. Recreational puzzle problems such as Sudoku and the N-Queens provide a good example of a constraint satisfaction problem since they can have multiple solutions.

## 6.1 Sudoku Puzzle

The Sudoku puzzle is played in a 9 by 9 grid made up of 3 by 3 sub grids containing one instance of digits from 1 to 9. The numbers are entered in such a way that each row, column and region (a 3* 3 sub grid) has exactly one instance of the digit ranging from 1 to 9.

There are digits that are already fixed for some cells, and they act as clues that restrict the solution space in such a way that there is only 1 correct way to populate the remaining regions.

## 6.1.1 Problem definition using Spreadsheet-specific functions

1. You would begin by setting the 9 by 9 grid, made up of 3 by 3 sub grids containing 1 instance of digits from 1 to 9. You would then proceed to add digits to certain cells, so that they act as clues that restrict the solution space, so that only 1 solution can be derived, as shown in the figure below:-

Fig 6.1 Sudoku problem Grid setup

2. You would then specify the variable range (A1:I9) using the spreadsheet-specific function **ssVarRanges(A1:I9)** as shown in bold in Fig 6.2.

3. You would then set the domain values min and max values of 1 to 9 by using the spreadsheet-specific function **ssDomain(A1:I9,1,9)**.

4. Then you would want to ensure that all the rows values in the 9 by 9 grid are different, so you would use the **ssRowsAllDifferent(A1:I9)**.

5. All the column values in the Sudoku grid would also have to be different, so you would use the ssColsAllDifferent(A1:I9).

6. The next thing would be to set each of the nine 3 by 3 regions values are different, so you would use the ssAllDifferent function on each of the nine 3 by 3 regions in the following way:

ssAllDifferent(A1:C3)
ssAllDifferent(A4:C6)
ssAllDifferent(A7:C9)
ssAllDifferent(D1:F3)
ssAllDifferent(D4:F6)
ssAllDifferent(D7:F9)
ssAllDifferent(G1:I3)
ssAllDifferent(G4:I6)
ssAllDifferent(G7:I9)

7. You would then proceed to specify the constraint ranges using the ssConstraintRanges function in the following manner ssConstraintRanges(A13:A24) as shown in Line 25. The **CellRange** A13:A24 contains the constraint specifications and exclude both the ssVarRanges and ssConstraintRanges functions. Figure 6.2 shows the Sudoku problem definition, before a solution is derived.

Fig 6.2 Sudoku problem definition

## 6.1.2 Solution using Spreadsheet-specific functions

1. Once the problem has been defined using spreadsheet-specific functions, you would select the "ParseBuild" sub-menu under the CSeE System menu as shown in the figure below.

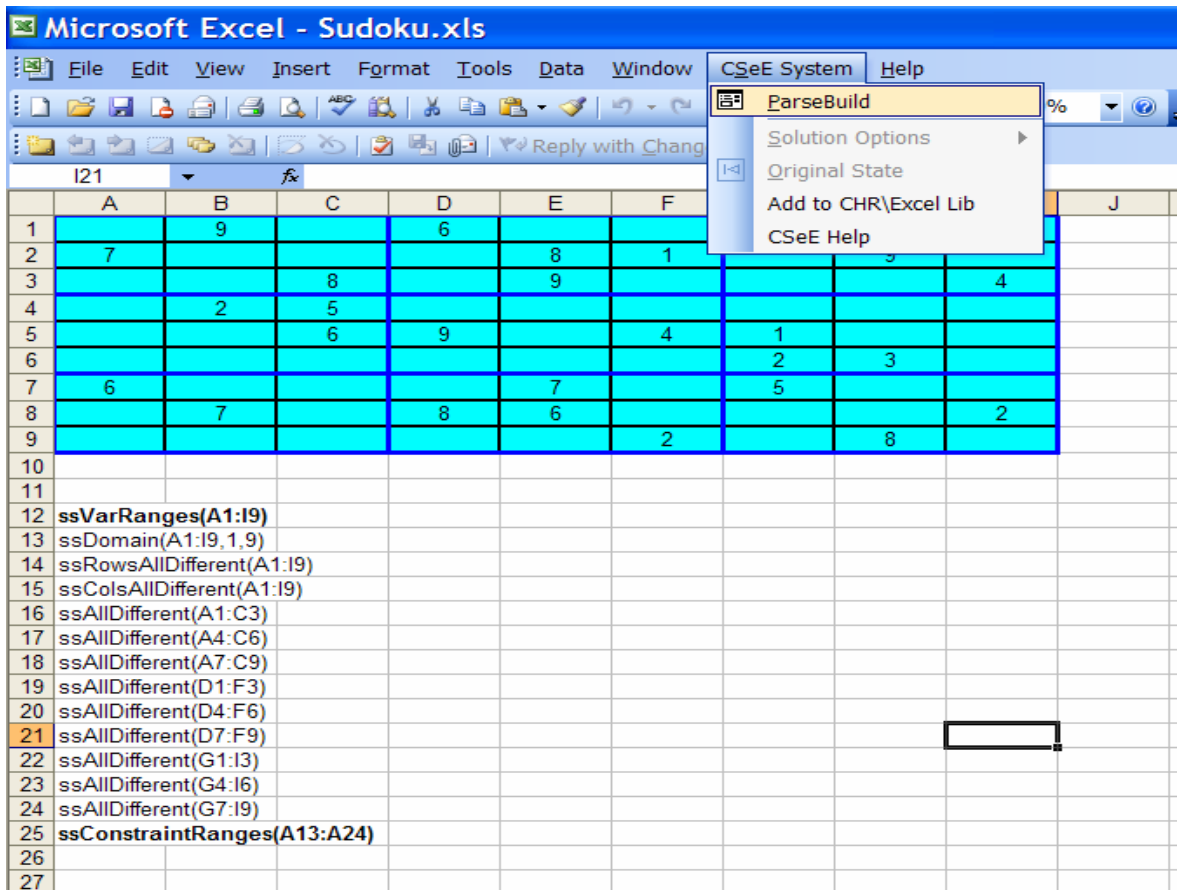

Fig 6.3 ParseBuild sub-menu selection

2. If there is a solution, the 1[st] solution would be displayed in the grid. And if there is more than one solution, the "Solution Option" sub menus would be enabled, along with the "Original State" menu as shown in the figure below.

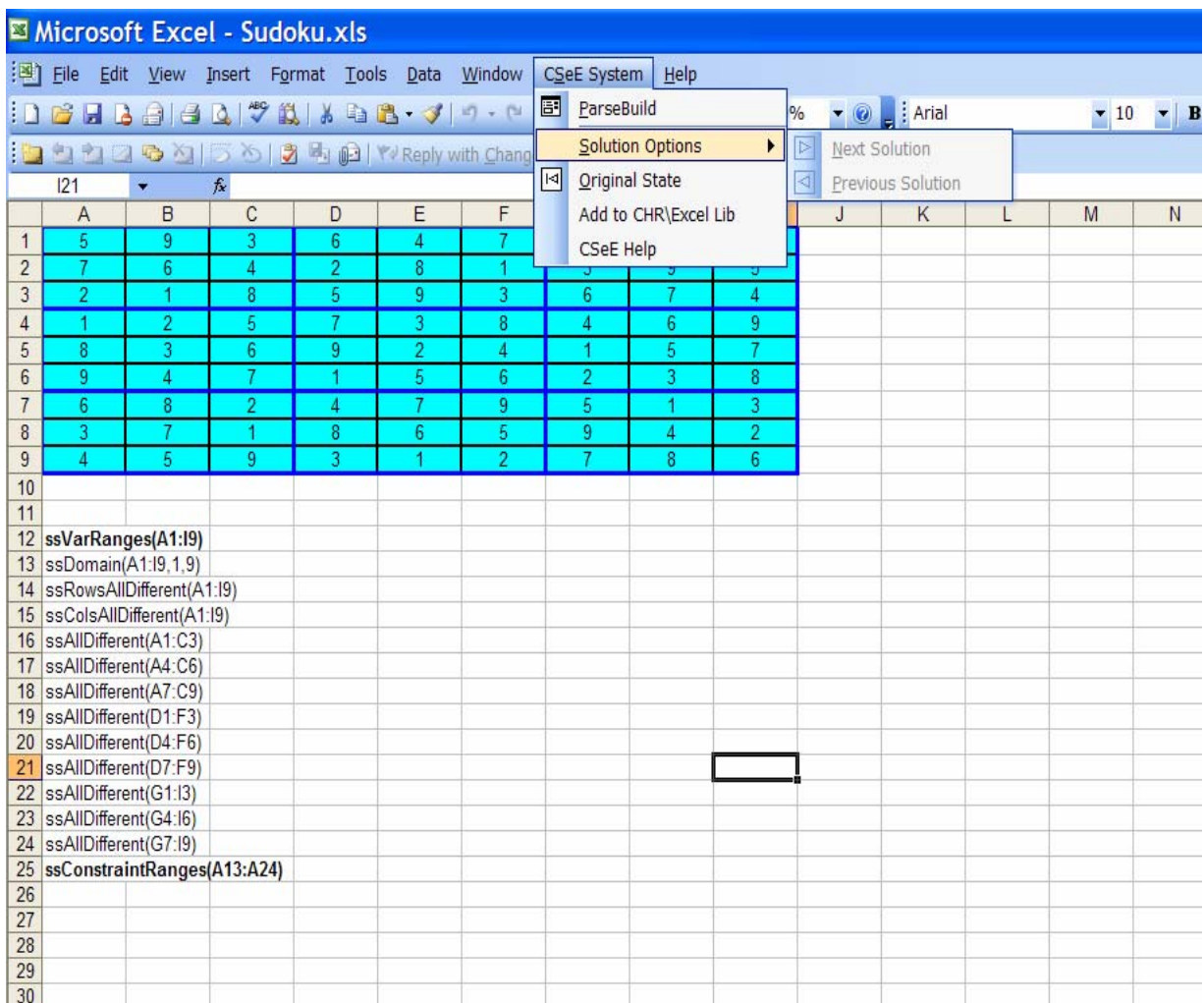

Fig 6.4 One Sudoku Solution with Solution Options sub menus disabled

3. Clicking on the "Original State" menu will return the grid to the original problem definition and also enable the "Solutions Options" sub menu "Next Solution". This would allow the user to view the solution derived before going back to the original state, so that they do not have to go through the process of re-parsing and re-building. The figure below shows the spreadsheet being returned to the original problem definition.

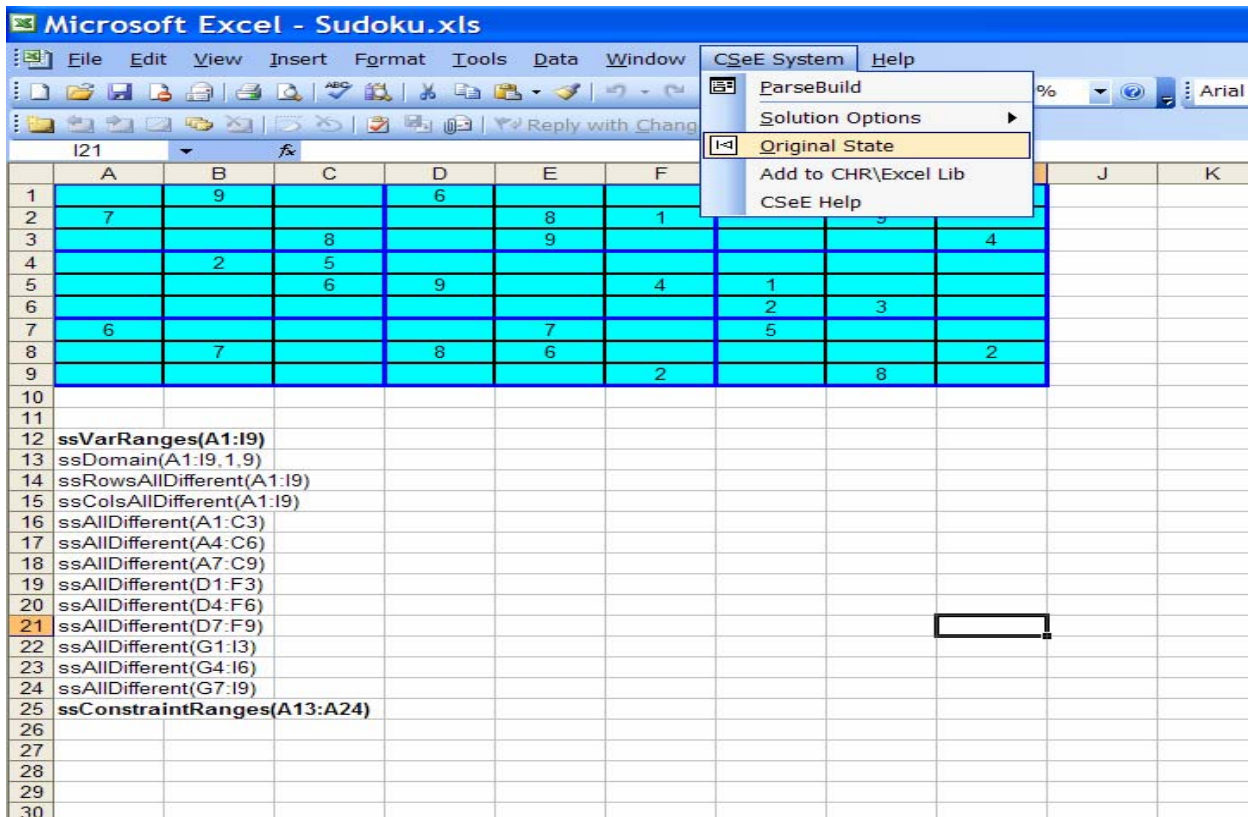

Fig 6.5 Original State of the problem definition

## 6.2 8-Queen Puzzle

The 8-queen puzzle deals with finding all the ways of placing 8 queens on an square 8 by 8 board, in such a way that no two queens are on the same row, column or diagonal. The solution to the 8 queen puzzle is a permutation in which for each column the row number where the queen appears is recorded.

### 6.2.1 Problem definition using Spreadsheet-specific functions

1. We begin by setting up an 8 by 8 grid, and use the insert function to specify the

variable ranges shown in the figure below. The figure shows the selection of the "CSeESystem.Functions" category and selecting the ssVarRanges function.

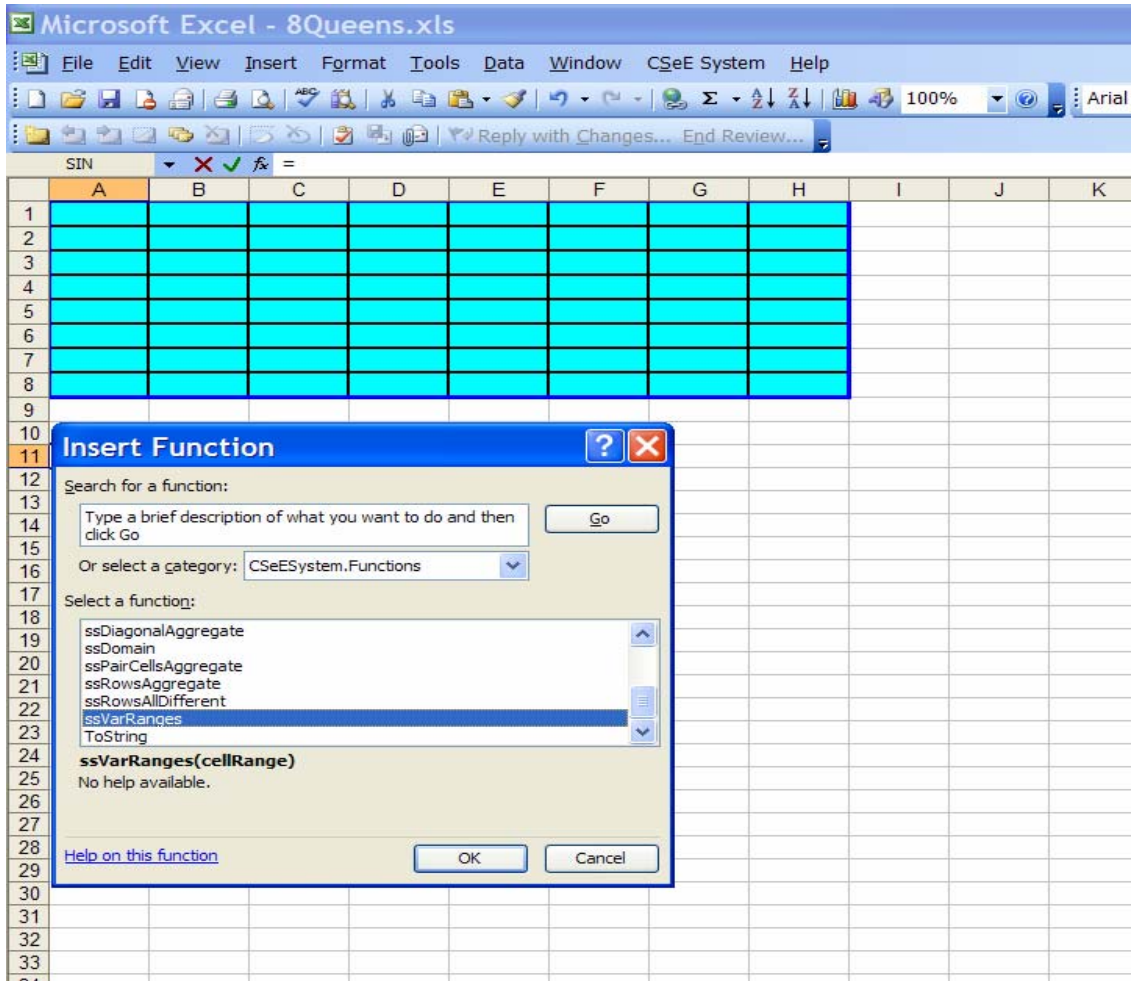

Fig 6.6 ssVarRanges insert function interface

2. The next figure shows the set function arguments interface with the ***CellRange*** A1:H8 being selected. Once the "OK" button is clicked, Line 11 will show the spreadsheet-specific function ssVarRanges(A1:H8), which as stated earlier will specify the variable range.

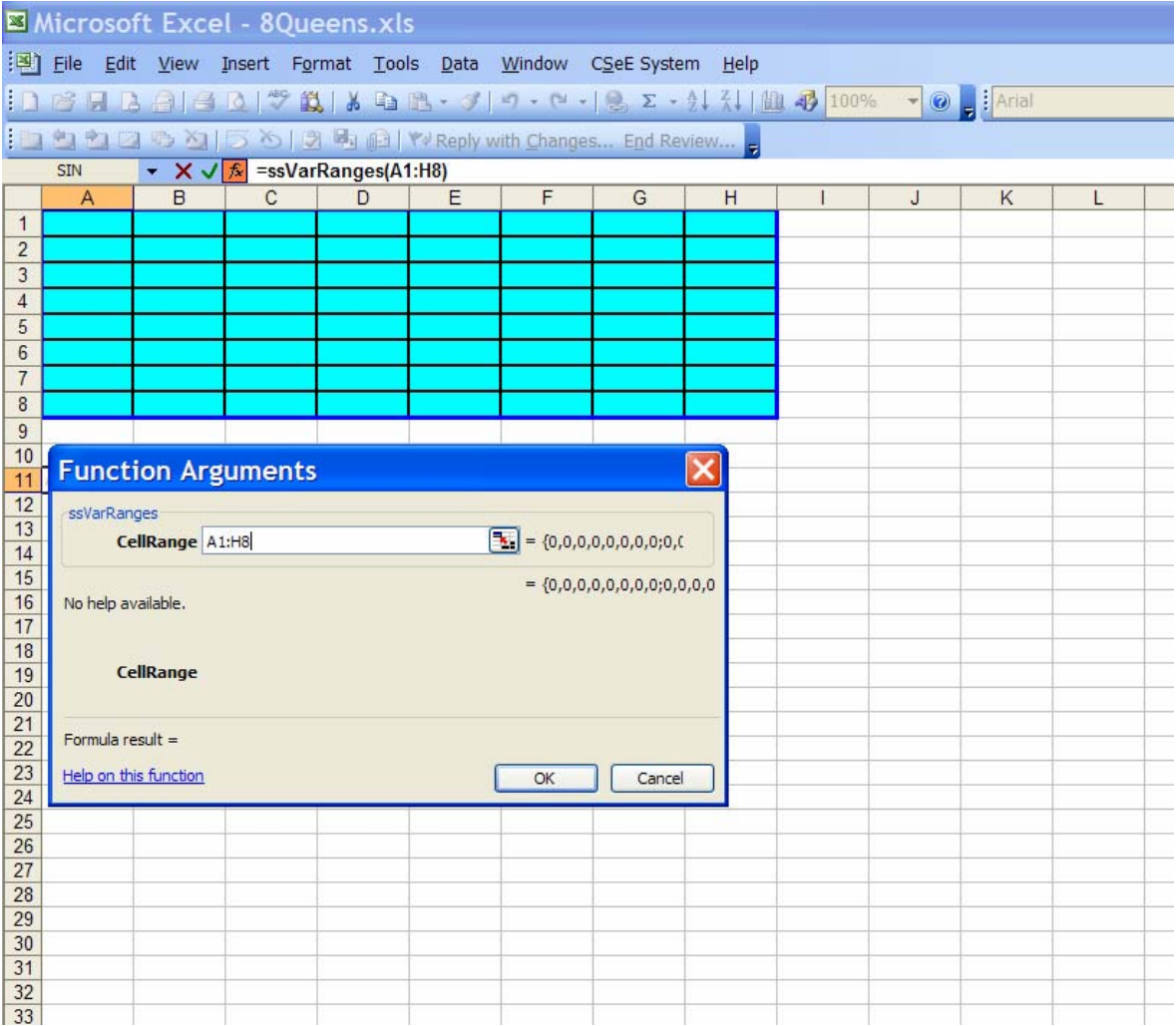

Fig 6.7 ssVarRanges set function argument interface

3. We need to state that the min and max values of the 8 by 8 grid will range from 0 to 1. We can do this by using the spreadsheet-specific function ssDomain through either the insert function interface or by just entering **ssDomain(A1:H8,0,1)**.

4. The next step would be to ensure that the sum of each row sums up to the value 1, by using the **ssRowsAggregate(+,A1:H8,#=,1)**.

5. We would then need to ensure that the sum of each column sums up to the value 1, by using the **ssColsAggregate(+,A1:H8,#=,1)**.

6. We would also need to ensure that the sum of each diagonal column from the top left corner to the bottom right corner of the grid and from the top right corner to the bottom left corner are less than or equal to the value 1, by using the 2 spreadsheet-specific functions below:

ssDiagonalAggregate(+,A1:H8,#=<,1)
ssBackDiagonalAggregate(+,A1:H8,#=<,1)

7. Next we would specify the constraint ranges by using the ssConstraintRanges(A12:A16) spreadsheet specific function, to derive the 8-queen problem definition shown in the figure below.

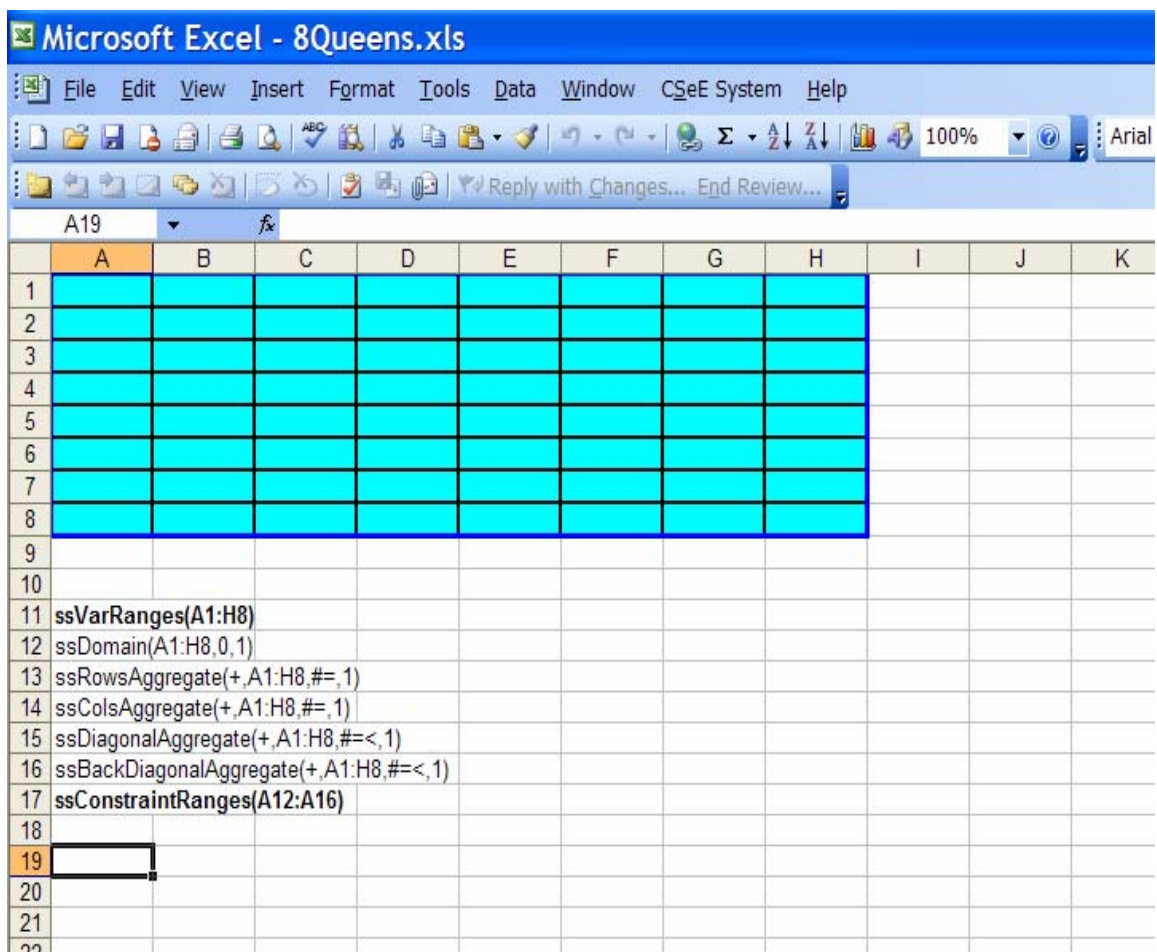

Fig 6.8 8-Queen problem definition

## 6.2.2 Solution using Spreadsheet-specific functions

1. We begin by clicking on the "CSeE System" menu's "ParseBuild" sub menu, which will parse and build the CSeE program. Additionally, it will display the derived 1st solution and enable the "Solution Option" sub menu buttons if there is more than 1 solution.

2. The figure below shows the 8-Queens problem has *92* solutions for a user to iterate through by clicking on the "Next Solution" and "Previous Solution" sub menus. You can return to the original problem definition by clicking on the "Original State" menu.

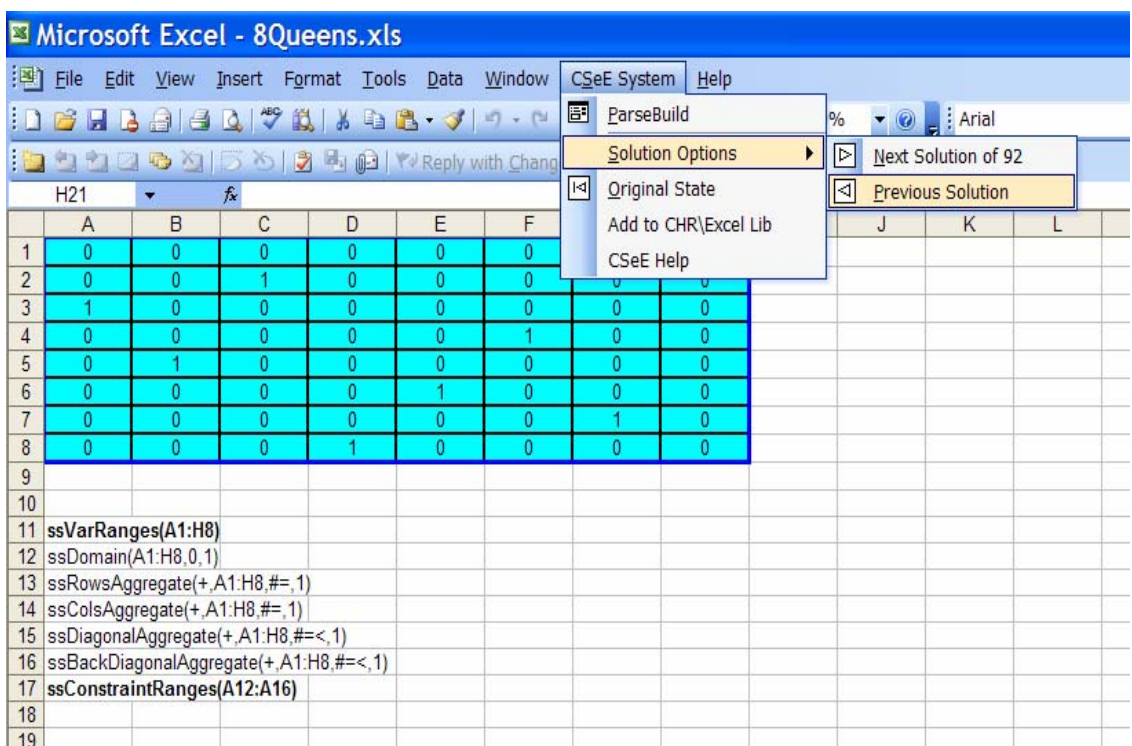

Fig 6.9 8-Queen solutions

# 6.3 Resource Allocation Example

The resource allocation for the table below seeks to *minimize* the total machine cost. Each machine has five possible jobs to choose from in order to minimizing the total cost all the machines.

## 6.3.1 Problem definition Spreadsheet-specific functions

1. We begin by setting up the problem definition space as shown in the figure below.

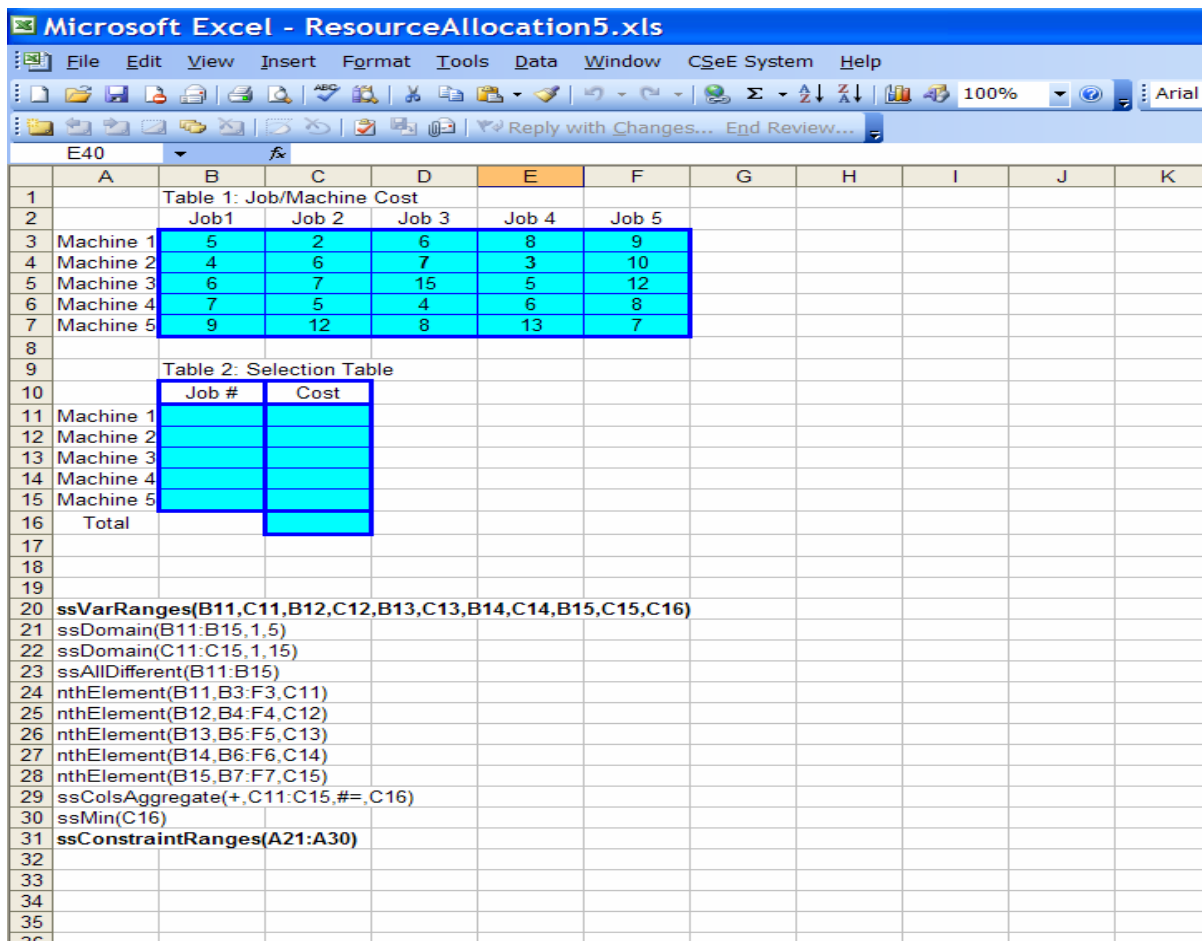

Fig 6.10 Resource Allocation problem

2. In order to include the Total cost value in the domain ranges in the selection table we use the ssVarRanges(B11,C11,B12,C12,B13,C13,B14,C14,B15,C15,B16).

3. We then specify the domain values for the jobs from 1 to 5 (since there are only 5 jobs), cost and mark that all the jobs have to be different by using the spreadsheet-specific functions shown below.

```
ssDomain(B11:B15,1,5)
ssDomain(C11:C15,1,15)
ssAllDifferent(B11:B15)
```

4. We then need to specify that the list of each of the selected jobs for each machine must be equal to the cost. In addition, we would need to specify that the sum of each of the machine's cost must be equal to the minimum total cost value. These requirements are stated by using the spreadsheet-specific functions below.

```
nthElement(B11,B3:F3,C11)
nthElement(B12,B4:F4,C12)
nthElement(B13,B5:F5,C13)
nthElement(B14,B6:F6,C14)
nthElement(B15,B7:F7,C15)
ssColsAggregate(+,C11:C15,#=,C16)
ssMin(C16)
ssConstraintRanges(A21:A30)
```

## 6.3.2 Solution using Spreadsheet-specific functions

1. We begin by clicking on the "CSeE System" menu's "ParseBuild" sub menu, which will parse and build the CSeE program. Additionally, it will display the derived 1[st] solution and enable the "Solution Options" sub menu buttons if there is more than 1 solution.

2. The figure below shows the Resource Allocation problem only has 1 solution and since there is only 1 solution, the "Solution Options" sub menus are disabled. You can return to the original problem definition by clicking on the "Original State" menu. Clicking on the

"Original State" menu, would also enable the "Solution Options" sub menu "Next

Solution" to allow users to go back to the derived solution, without having to re-parse and re-build the solution.

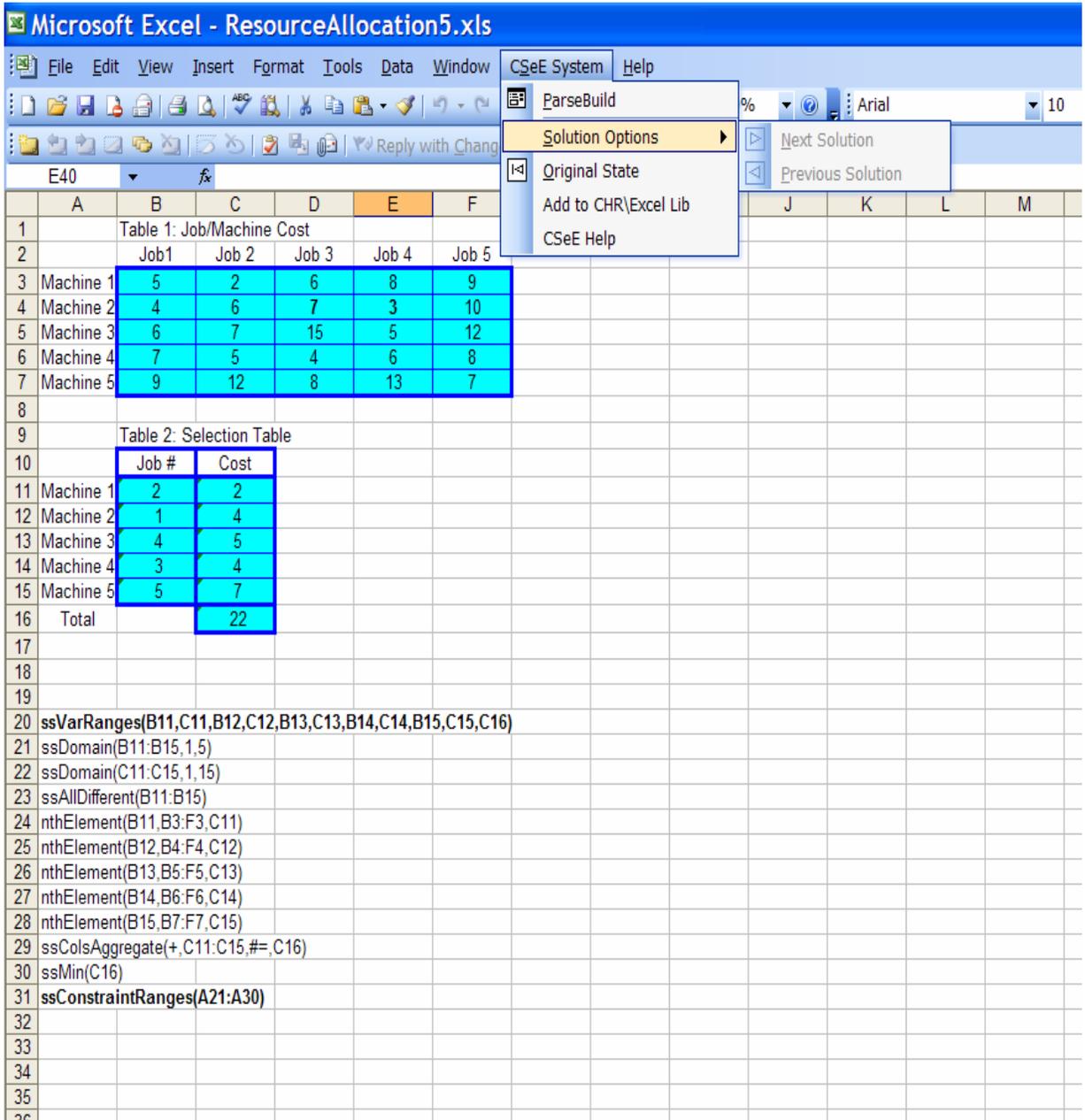

Fig 6.11 Resource Allocation solution

# 6.4 Adding to the CHR\Excel Library

In order to add constraint logic predicates to the "excel.pl" library, we would go to the "CSeE System" menu and choose the "Add to CHR\Excel Lib" sub menu. The interface in the figure below will show and display the "excel.pl"'s path and contents

## 6.4.1 Adding to the Excel Library

1. The desired prolog predicate can then be added to the "excel.pl" file as shown by the highlighted portion of the figure below.

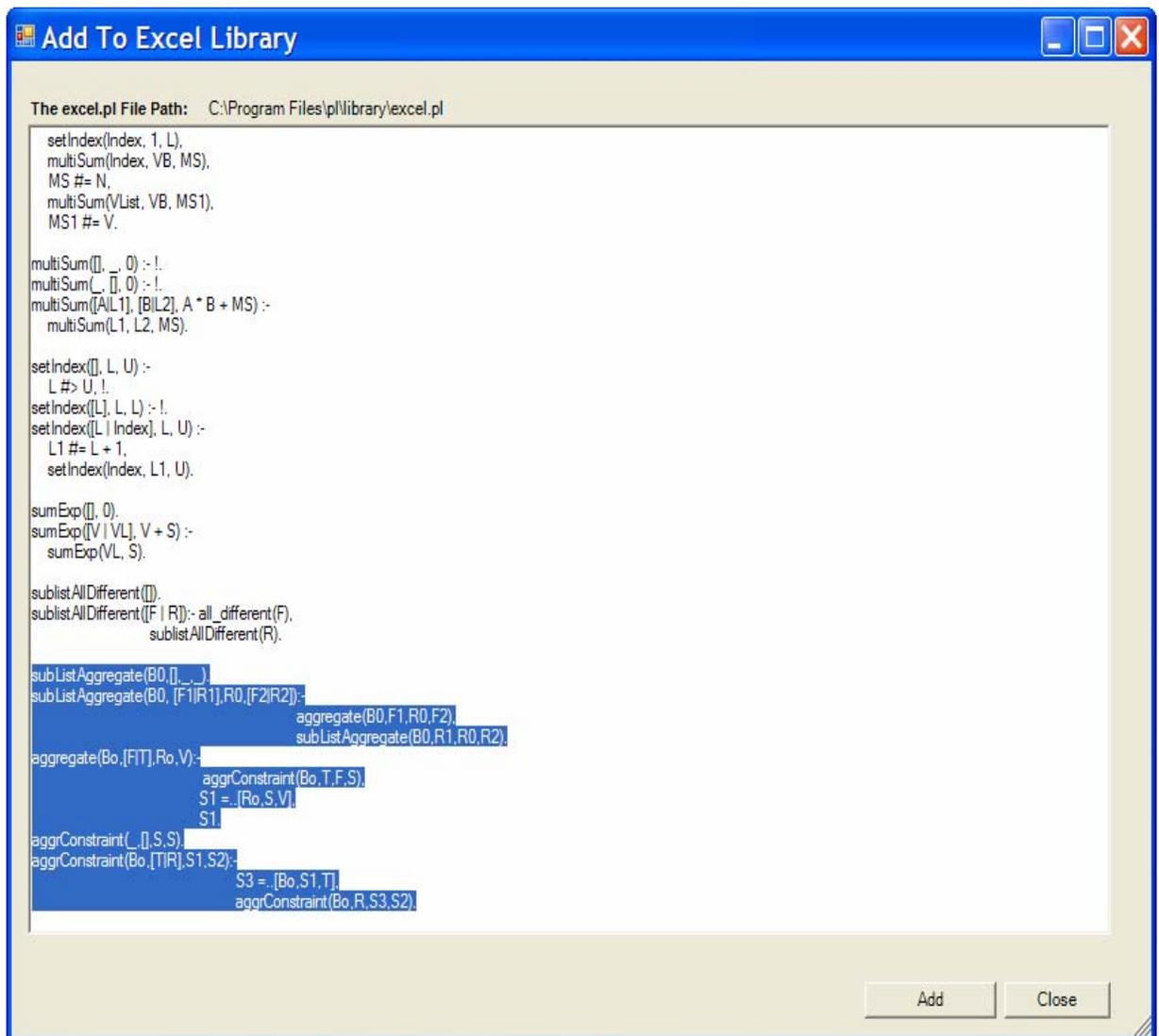

Fig 6.12 Adding a new prolog predicate

2. The prolog predicate that is going to be called from an external file would need to be added to the module section as shown by the highlighted portion of the figure below:

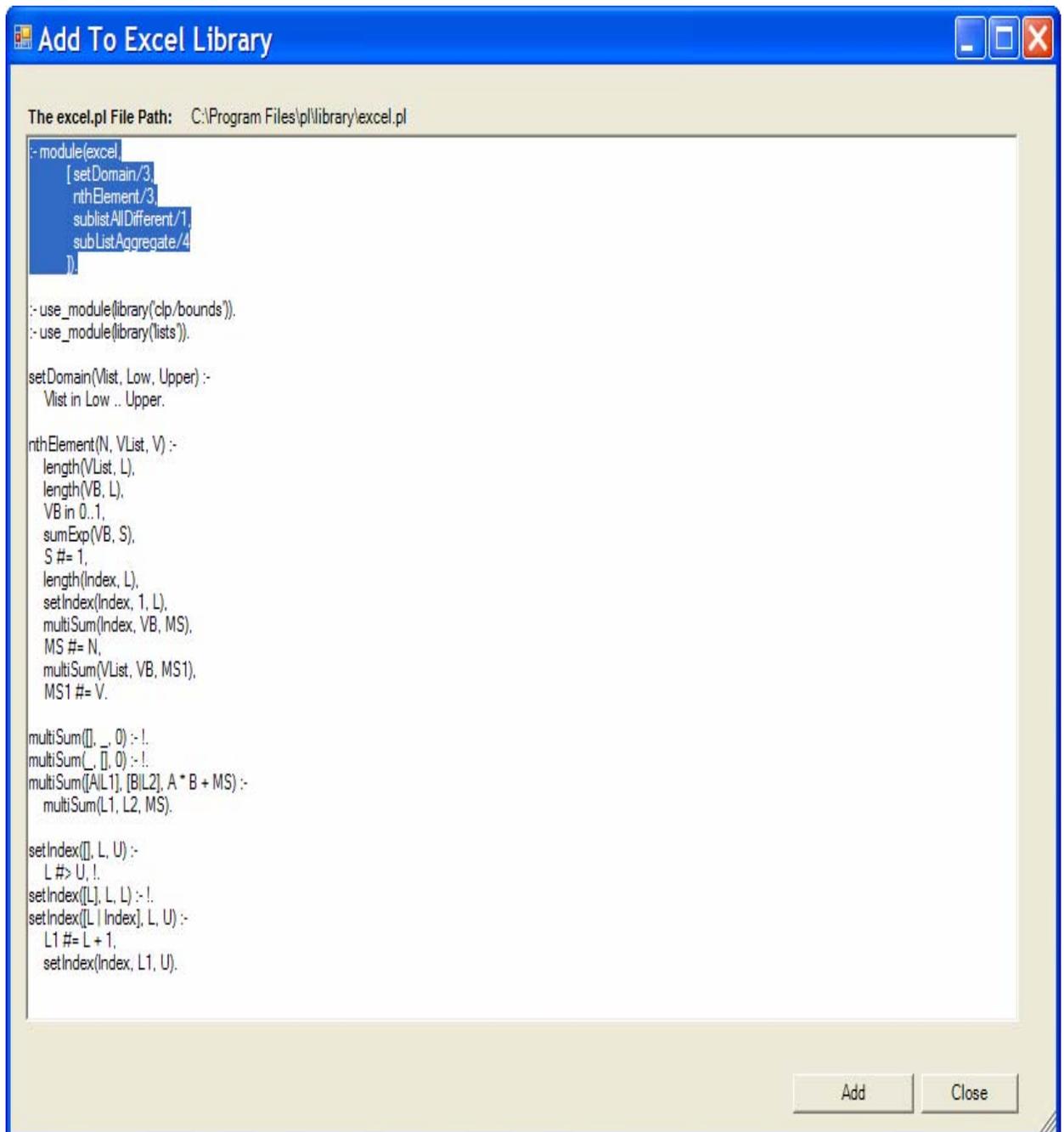

Fig 6.13 Adding the new prolog predicate to the module

3. We then append the predicate to the library by clicking on the "Add" button, which will cause a backup to be created under the following path `"C:\Program Files\pl\MyCLPAddIN\excel.pl"`. It would then proceed to overwrite the excel.pl file and upon completion a message "Appending to excel.pl completed…" message will be displayed. Once that completed message is received, we can then close the interface by clicking on the "Close" button as shown in the figure below.

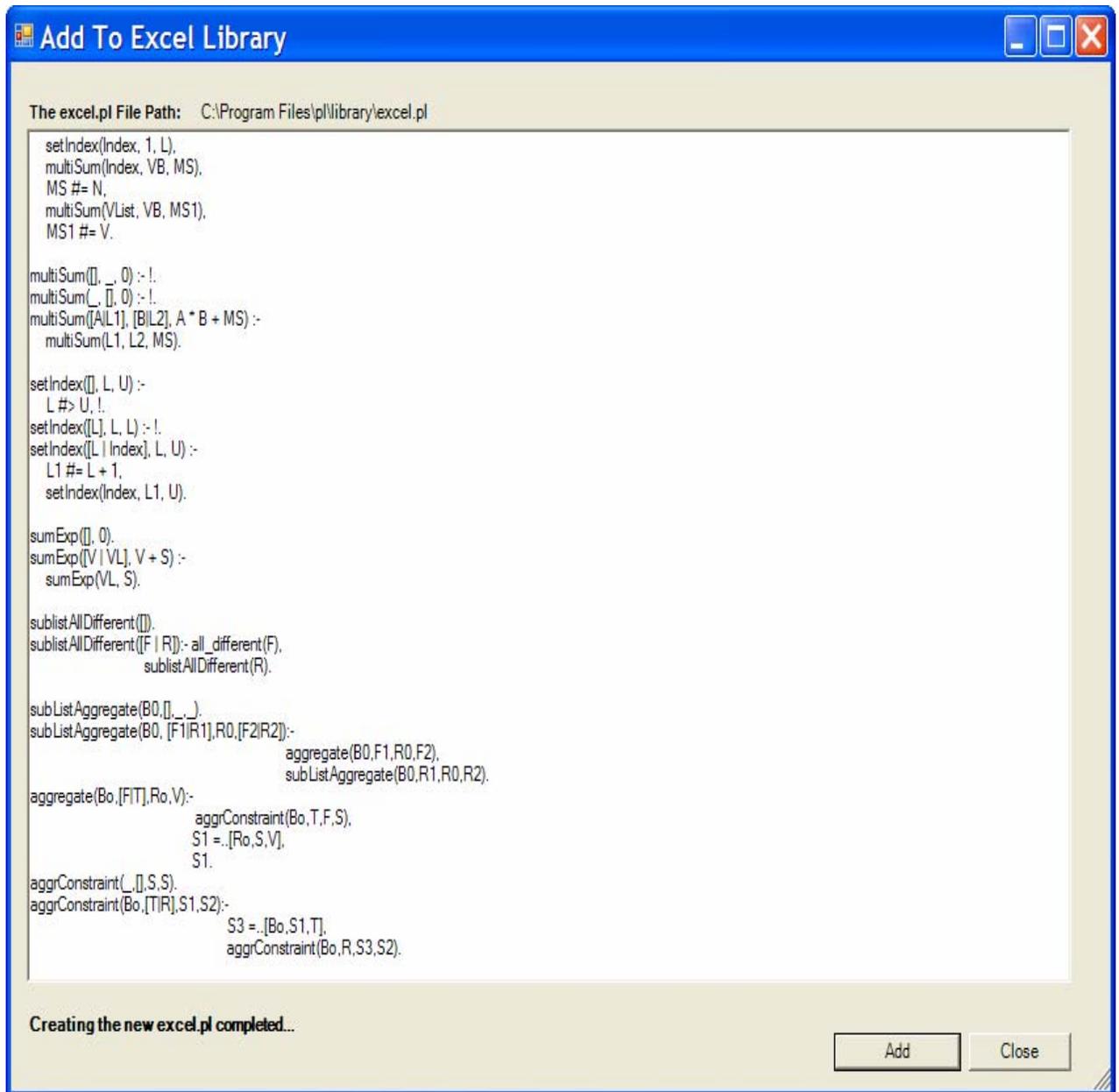

Fig 6.14 Add to Excel Library interface

## 6.5 CSeE Help

In order to access a help menu about the CSeE system's syntax and semantics, we would go to the "CSeE System" menu and choose the "CSeE Help" sub menu as shown in the figure below.

Fig 6.15 CSeE Help interface

# CHAPTER 7

# Conclusion and Future Extensions

## 7.1 Conclusion

Excel Spreadsheet users encounter constraint satisfaction problems on a daily basis. As such the CSeE system is a framework incorporating the spreadsheet paradigm with a Constraint Logic Programming (FD) solver so that it becomes a useful and indispensable tool for the users. The CSeE's seamless integration with the Excel and SWI-Prolog, along with its spreadsheet-specific language extends the spreadsheet usability.

The CSeE system is the first time (as known) that a spreadsheet-specific language allows users to specify constraints that exist among the cell in a declarative and scalable way, so as to solve constraint satisfaction problems. The framework significantly simplifies the complexity involved in developing solutions to many constraint-based application domains, from within the Excel's spreadsheet interface.

## 7.2 Future Extensions

Given constraint satisfaction problems encompass a wide array of application domains; the spreadsheet-specific library can be further enhanced and optimized as solutions for these different application domains are discovered. In addition, the spreadsheet-specific library could provide a different perspective and methodology for users when they attempt to explore new ways of solving many their constraint satisfaction problems. Some proposed future extensions of the CSeE system are listed below:

*1. Making the system asynchronous, instead of synchronous:*

The CSeE system 's interface with the .NET wrapper system uses the synchronous methodology, which means that once it sends a request to the SWI-Prolog system, it waits until a result is received. Making the system asynchronous, would allow other solutions to be derived and queued while the 1$^{st}$ solution is being derived, thereby increasing productivity.

*2. Making the system web accessible:*

Making it web accessible through either a Web Service or a web-site would allow for it to become a collaboration tool, while making updates and maintenance more manageable.

*3. Increasing the compatibility with existing Excel functions:*

Integrating the existing Excel functions with the spreadsheet-specific function would increase the reusability of the CSeE system.

*4. Large-scale constraint problem solving:*

Enhancing the system's ability to solve large scale problems, through the use of custom designed propagation rules and consistency techniques, would make the system an invaluable tool.

*5. Creating spreadsheet-specific languages for different application-domains*

More spreadsheet-specific languages can be developed for solving different application domains such as Operation Research problems, Hardware/Software verification and design, biological problems etc…